\ifdefined\pdfminorversion\pdfminorversion=7\fi
\documentclass[11pt]{article}

\usepackage[letterpaper,margin=1in]{geometry}
\usepackage[T1]{fontenc}
\usepackage[utf8]{inputenc}
\IfFileExists{newtxtext.sty}{\usepackage{newtxtext,newtxmath}}{\usepackage{mathptmx}}
\usepackage[final]{microtype}
\usepackage{setspace}
\usepackage{amsmath}
\usepackage{bm}
\usepackage{mathtools}

\usepackage{graphicx}
\usepackage{tikz}
\graphicspath{{figures/}}
\usepackage{booktabs}
\usepackage{multirow}
\usepackage{array}
\usepackage{float}
\usepackage[section,below]{placeins} 
\usepackage[font=small,labelfont=bf,labelsep=period,skip=6pt]{caption}

\usepackage[title,titletoc]{appendix}

\usepackage{authblk}

\usepackage[numbers,square,comma,sort&compress]{natbib}
\AtBeginDocument{\renewcommand{\natexlab}[1]{}} 

\usepackage{xcolor}
\usepackage[colorlinks=true,linkcolor=blue!60!black,citecolor=blue!60!black,
            urlcolor=blue!60!black,breaklinks=true]{hyperref}
\usepackage[capitalise,nameinlink,noabbrev]{cleveref}
\crefname{equation}{Eq.}{Eqs.}
\Crefname{equation}{Equation}{Equations}
\crefname{appendix}{Appendix}{Appendices}
\crefname{subappendix}{Appendix}{Appendices}
\creflabelformat{equation}{(#2#1#3)}

\newcommand{\KDF}{\mathrm{KDF}}                 
\newcommand{\KDFd}{\overline{\mathrm{KDF}}}     
\newcommand{\bbeta}{\bar{\beta}}                
\newcommand{\x}{\mathbf{x}}
\newcommand{\z}{\mathbf{z}}
\newcommand{\y}{\mathbf{y}}
\newcommand{\X}{\mathbf{X}}
\newcommand{\K}{\mathbf{K}}
\newcommand{\I}{\mathbf{I}}
\newcommand{\kv}{\mathbf{k}}
\newcommand{\bphi}{\bm{\phi}}
\newcommand{\bPhi}{\bm{\Phi}}
\newcommand{\bbet}{\bm{\beta}}
\newcommand{\GP}{\mathcal{GP}}
\newcommand{\Normal}{\mathcal{N}}
\newcommand{\Rsq}{R^{2}}
\newcommand{\T}{\mathsf{T}}

\title{\textbf{A Hybrid Gaussian Process--Extreme Gradient Boosting Machine Learning Framework for Robust Knockdown Factor Prediction and Reliability-Based Design of Thin Spherical Shell Caps}}

\author[1]{Budhaditya De}
\author[2,*]{Rohan Majumder}
\author[1]{Devasmit Dutta}
\author[2]{Aman Deep Gupta}
\author[3]{Sudib Kumar Mishra}
\affil[1]{Department of Civil and Environmental Engineering, University of California, Los Angeles, CA, USA}
\affil[2]{Faculty of Engineering Science and Technology, Adani University, Ahmedabad, Gujarat, India}
\affil[3]{Department of Civil Engineering, Indian Institute of Technology Kanpur, Uttar Pradesh, India}
\affil[*]{Corresponding author: \href{mailto:rohanmajumder1989@gmail.com}{rohanmajumder1989@gmail.com}}
\date{}

\begin{document}
\maketitle

\begin{abstract}
Elastic buckling remains a critical concern in the design of deep spherical shell caps, given that experiments reveal a significant reduction of critical loads from those predicted by linearized eigen-buckling analysis. This discrepancy has been attributed to initial geometric imperfections and is often reconciled through a dimensionless parameter, namely the knockdown factor (KDF). However, conventional KDFs (NASA SP-8007; Eurocode~3) are highly conservative, as they represent the lower bound of an extensive body of experimental data. This study aims to narrow the conservatism in the KDFs for spherical shell caps through machine learning (ML) regression models trained on experimental data on spherical shell buckling. A hybrid model is proposed by combining Gaussian Process Regression (GPR) with Extreme Gradient Boosting (XGBoost), and its improved performance is demonstrated over conventional ML algorithms, i.e., the Artificial Neural Network (ANN), Support Vector Regressor (SVR), Random Forest Regressor (RFR), and Histogram-based Gradient Boosting (HGBoost). Each of these models uses six input features and a single output, i.e., the KDF. The most critical feature affecting the KDF is ascertained through a sensitivity analysis. The study concludes with a reliability-based design framework that relies on the predictive ML model to obtain safety-consistent, yet less conservative, KDFs.
\end{abstract}


\section{Introduction}
\label{sec:intro}

Deep spherical shells are widely used in engineering due to their excellent load-carrying capacity, high strength-to-weight ratio, and balanced membrane--bending action derived from their curvilinear geometry. Domed shell caps are employed as end closures in cylindrical pressure vessels; alternative geometries such as hemispheres, tori-spheres, ellipsoids, and tori-cones are also used for the same purpose. Spherical shells are further used in lattice domes, actuators, carbon nanostructures, and bio-encapsulation. Pollen grains and viruses are common biological forms of spherical nature~\citep{krivoshapko2007ellipsoidal,wang2018unified,nguyen2005elasticity}. Despite their advantages, the buckling and post-buckling behaviour of spherical shells under external pressure and of cylindrical shells under axial compression is highly sensitive to geometric imperfections. Post-buckling load drops of 70--80\%, relative to theoretical predictions, are commonly observed~\citep{simitses1986buckling,bisagni2003experimental}. The extensive experimental campaign of NASA SP-8007~\citep{weingarten1968sp8007} and accompanying theoretical studies modelled such imperfections as dimples or sinusoidal shapes. This disparity between experimental and theoretical (linearized eigen-analysis) estimates leads to the adoption of the knockdown factor (KDF) in design. The KDF multiplied by the linearized buckling load offers a conservative (lower-bound) design against post-critical drops. The interaction between closely spaced buckling modes, modal coupling due to geometric nonlinearity, and modal multiplicity are key to such load drops.

Depending on their radius-to-thickness ratio ($R/t$), shells are classified as either ``thin'' or ``thick''. Thick shells show pronounced out-of-plane shear deformation. \citet{eggwertz1990buckling} classified spherical shells as ``shallow'' or ``deep'' caps based on the central angle ($\phi$) subtended at the centre of curvature by their arc. Increasing curvature generally stabilizes the shell, but deep shells, despite this curvature, are prone to lower KDFs and greater imperfection sensitivity. Pioneering research by von K\'arm\'an, Tsien, and Koiter~\citep{vonkarman1940influence,tsien1942theory,koiter1956buckling} attributed the load drops to nonlinear modal interaction and to the erosion of potential energy by imperfections. Investigations by Hutchinson~\citep{hutchinson2018imperfections}, the NASA SP-8032 recommendations~\citep{nasa1969sp8032}, and subsequent evaluations by \citet{singer2002buckling} and \citet{domb2002nonlinear} also substantiate this claim. While KDF-based design, though conservative, remains the standard, recent developments in stochastic mechanics, Monte Carlo simulation, and probabilistic design have acknowledged the uncertain nature of imperfections~\citep{schenk2003buckling}.

The Energy Barrier Criterion (EBC) framework offers an alternative avenue for studying shell buckling and imperfection sensitivity by relating post-critical stability to an ``energy landscape'' whose valleys and ridges define metastability~\citep{hutchinson2018imperfections}. This approach has been applied to assess robustness against both static and dynamic perturbations, with probabilistic evaluations quantifying the scatter in the KDFs. The reduced stiffness method, developed by Croll and co-workers~\citep{croll1995towards}, provides an alternative means to estimate lower-bound buckling loads by modelling the stiffness degradation caused by imperfections. This approach has proven useful in capturing highly nonlinear responses and in setting imperfection tolerances for desired performance levels. The stabilizing influence of boundary conditions, the role of curvature, and the susceptibility of thick shells to elastoplastic buckling have also been noted. Additionally, spherical caps can display complex dynamic responses (including chaos), necessitating nonlinear dynamic stability assessments~\citep{sahu2007research,rizzetto2019nonlinear}. Imperfections are conveniently modelled as zero-mean, stationary, isotropic Gaussian random fields, characterized by a root-mean-square amplitude and power spectral density~\citep{papadopoulos2005effect}, although non-Gaussian models~\citep{papadopoulos2009buckling} have also been proposed to better capture manufacturing variability. Over-conservatism in KDFs can be further reconciled through Reliability-Based Design (RBD), which accounts for the statistical margin of safety~\citep{hao2015hybrid}. Nonlinear stability analyses use the Riks algorithm to capture post-buckling behaviour, including abrupt mode localization. Coupled nonlinear stochastic analysis offers a statistically grounded design tool that directly connects imperfection characteristics, geometric parameters, and reliability targets ($\bbeta$).

Metamodelling methods, such as RS approximation and Kriging, have traditionally been employed to estimate nonlinear buckling loads as functions of design parameters~\citep{mukhopadhyay2017critical}. However, these approaches struggle to capture highly nonlinear relationships. Considering random imperfections is critical for accurate buckling load estimation. While Monte Carlo simulation can improve accuracy, its computational cost remains high. The prediction of KDFs for shell buckling has traditionally relied on experimental data, complementary simulations, statistical methods, and lower-bound estimates. However, the complexity of the datasets and the absence of comprehensive theoretical models have led to the growing adoption of machine learning (ML) approaches, for instance Artificial Neural Networks (ANNs)~\citep{mandal2021application}. The ANN mimics human neural functionality and excels at identifying patterns in high-dimensional data that are otherwise challenging for conventional statistical or mathematical models. ANNs consist of interconnected units (neurons) that capture complex input--output relationships. The ANN architecture may include single or multiple layers, with variations in signal flow (feedforward or recurrent) and node counts, enabling flexible modelling. Multi-layered feedforward ANNs have been successfully applied to predict the buckling load of laminated composite stiffened panels under in-plane shear~\citep{mallela2006buckling}. Hybrid ANN--finite element (FE) models have also been employed by \citet{kumar2019stochastic} to study higher-mode effects on sandwich plates. Other notable contributions include critical load prediction for cylindrical shells using back-propagation ANNs~\citep{waszczyszyn2002neural}, polyvinyl chloride (PVC) cylindrical shells~\citep{markopoulos2007prediction}, and composite cylindrical shells via ANN--Support Vector Regression (SVR) models~\citep{pitton2019buckling}. ANN approaches have been further refined by incorporating physical constraints, as in physics-informed ANNs (PINNs) for axially compressed thin cylindrical shells~\citep{tao2020physics} and, more recently, for robust buckling load prediction and reliability-based design of thin truncated conical shells~\citep{dutta2026physics}. Studies have extended beyond shells, covering composite panels~\citep{malik2013ann}, steel arch-supported shells~\citep{hasanzadehshooiili2012neural}, thin rectangular plates with welding imperfections~\citep{sadovsky2011artificial}, and columns~\citep{mukherjee1996prediction}. The integration of ANN-based methods represents a significant advancement in predicting KDFs and buckling loads, offering higher accuracy and adaptability than traditional methods.

Recent advances in ML have also enabled improved modelling of highly nonlinear mappings. Support Vector Machines (SVMs) have been used to bypass iterative optimization processes; hybrid approaches, such as SVM with simulated annealing, were shown to reduce computation time by as much as 50\%~\citep{tamura2018machine}. Random Forest Regression (RFR) models have also been applied to predict structural deformation~\citep{dai2018statistical}. A key limitation of ML models is their interpretability: trained models often function as black boxes, making them difficult to integrate directly into the design process. Genetic programming can produce explicit regression expressions~\citep{arnaldo2014multiple}; however, generating such models for complex systems remains challenging. Binary decision trees, including Random Forests, can theoretically offer transparent models. Ongoing research addresses ANN and deep neural network (DNN) interpretability, but complex rules or closed-form functions remain impractical for use in design. SVR with a linear kernel can yield explicit functions at the expense of predictive accuracy for highly nonlinear problems.

The review indicates that although data-driven techniques have been utilized for buckling prediction and stability analysis of plates and shells (cylindrical and conical geometries), including the robust ML-based prediction and sensitivity assessment of KDFs in truncated conical shells~\citep{majumder2026robust}, their application to spherical (synclastic) shells is scanty. Following the approach of \citet{wagner2018robust}, appropriate design variables (DVs), involving shell geometry and material properties, are adopted as the input features, with the KDF as the sole output. The training dataset is obtained from available experimental data~\citep{wagner2018robust}. Alternative models, namely ANN, SVR, RFR, Histogram-based Gradient Boosting (HGBoost), and a hybrid Gaussian Process Regression--Extreme Gradient Boosting model (GPR--XGB), are trained, tested, and validated on this dataset to predict the KDFs of the spherical shell cap for possible combinations of the input features. The performance of the alternative models is verified using the coefficient of determination ($\Rsq$) and root-mean-squared error (RMSE) on the test set, followed by a sensitivity analysis that determines the most influential input feature affecting the predicted KDF. Building on the hybrid Gaussian Process--XGBoost framework recently developed by the authors for assessing buckling strength reduction in truncated conical shells~\citep{majumder2026data}, and unlike conventional predictive ML models that directly approximate the input--output relationship, the hybrid GPR--XGB framework decomposes the prediction into a global probabilistic trend and a local residual correction. GPR captures the underlying smooth physics of shell buckling while providing uncertainty estimates, whereas XGBoost learns from the remaining nonlinear residuals arising from imperfection-sensitive behaviour. This complementary architecture improves predictive accuracy and robustness when learning from sparse experimental datasets, and it additionally provides predictive uncertainty.

Finally, an RBD procedure is presented that uses the predictive KDF model to reduce conservatism. The RBD enables efficient repeated evaluations of the limit state function (LSF), bypassing tedious FE analyses, and yields the shell design variables required to achieve a target reliability index ($\bbeta$).

\section{FE Modelling of the Shell Geometry}
\label{sec:fe}

\begin{figure}[!htbp]
  \centering
  \includegraphics[width=0.55\textwidth]{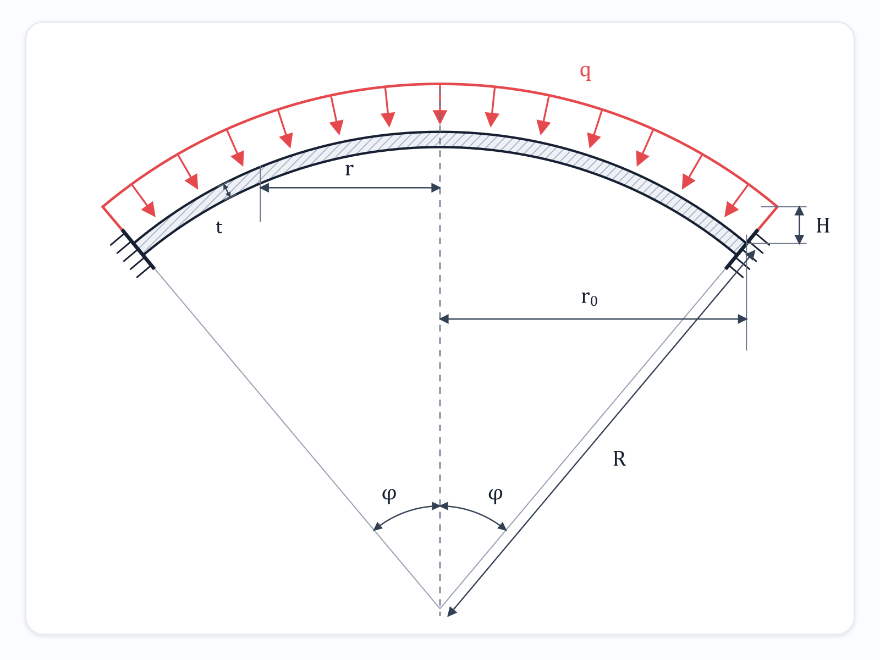}
  \caption{Spherical shell cap (adapted from \citep{majumder2024reliability}) subjected to uniform external pressure.}
  \label{fig:shell_cap}
\end{figure}

The shell is made of aluminium (Al-6061-T6) with a radius ($R$) of 1000~mm, thickness ($t$) of 2.5~mm, and a meridional angle ($\phi$) of $60^{\circ}$~\citep{majumder2024reliability}, as shown in \cref{fig:shell_cap}.

\begin{figure}[!htbp]
  \centering
  \includegraphics[width=\textwidth]{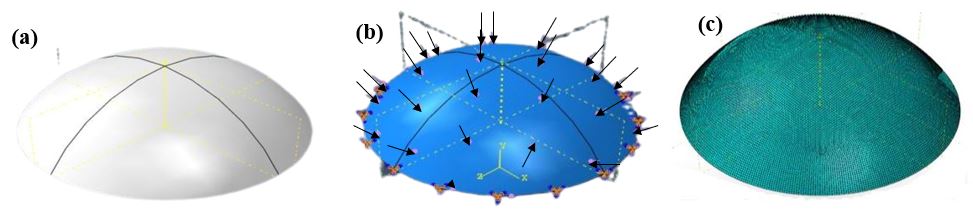}
  \caption{Spherical shell: (a) geometry; (b) uniform external pressure with clamped boundary condition; and (c) FE mesh \citep{majumder2024reliability}.}
  \label{fig:fe_model}
\end{figure}

The shell cap is characterized by a density ($\rho$) of 2700~kg/m$^{3}$, an elastic modulus ($E$) of 68.9~GPa, a Poisson's ratio ($\nu$) of 0.33, and a yield stress ($Y$) of 240~MPa. Within the examined loading regime, the material remains linearly elastic. The cap is modelled with clamped boundary conditions along its periphery. With the adopted geometry, $R/t$ equals 400 (lying between 100 and 3000), which is consistent with the NASA SP-8032 guidelines~\citep{nasa1969sp8032}; exceeding the threshold of 250, the shell cap is classified as ``thin''. The other geometric parameter, $\alpha = \sqrt{2\phi\,R/t}$, equals 28.94, which is greater than 7; hence, the shell cap is ``deep''~\citep{eggwertz1990buckling}. The ratio of dome height to base radius ($H/r_0$) equals 0.58, which lies in the range 0.035--1. Collectively, these metrics confirm that the shell cap is deep and thin. Furthermore, the NASA-recommended geometric shell (shape) parameter, $\lambda = \bigl[12(1-\nu^{2})\bigr]^{1/4}\sqrt{R/t}\,\bigl(2\sin(\phi/2)\bigr)$, is calculated as 36.17 (falling within the range 2--70), consistent with the criteria for stability assessment, as the KDF for spherical shells is expressed as a function of $\lambda$~\citep{wagner2018robust}. The spherical shell geometry, applied loading, boundary conditions, and FE mesh (in ABAQUS) are illustrated in \cref{fig:fe_model}(a)--(c).

The shell model is discretized with fully integrated four-noded shell elements (S4). A mesh convergence study was conducted with respect to the lowest critical pressure to determine the appropriate element size. The uniform pressure corresponding to yielding of the shell was found to be 1.20~MPa. The fundamental critical (elastic) pressure ($q_c^{1}$) following Zoelly and van der Neut~\citep{vanderneut1932elastische},
\begin{equation}
  q_c^{1} = \frac{2E}{\sqrt{3(1-\nu^{2})}}\left(\frac{t}{R}\right)^{2},
  \label{eq:zoelly}
\end{equation}
is 0.53~MPa, which is much lower than the yielding pressure of 1.20~MPa; the shell is therefore expected to fail by elastic buckling. A summary of the linear and nonlinear buckling analyses is presented in \cref{app:fe}.

\section{Machine-Learning-Based Knockdown Factor Prediction}
\label{sec:ml}

Alternative ML predictive models are considered with six input features, namely $R/t$, $H/r_0$, $\lambda$, $Y/E$, $\nu$, and the fabrication imperfection (quality) parameter $Q$. The KDF is treated as the sole output for a spherical shell cap with clamped (fixed) boundary conditions (BC) along its periphery, subjected to uniform external pressure. The entire experimental dataset is obtained from~\citet{wagner2018robust} and split into a training set and a test set. The training set comprises 80\% of the dataset, whereas the test set consists of the remaining 20\%. A total of 582 data points (experimental observations) are considered. A statistical summary of the dataset used for training the ML models is provided in \cref{tab:data}.

\begin{table}[!htbp]
  \centering
  \caption{Statistical summary of the data used to develop the ML predictive models.}
  \label{tab:data}
  \begin{tabular}{lrrrr}
    \toprule
    Parameter & Minimum & Maximum & Mean & Std.\ deviation \\
    \midrule
    $R/t$     & 10.031 & 4345.745 & 778.519  & 651.191  \\
    $H/r_0$   & 0.023  & 1.000    & 0.246    & 0.283    \\
    $\lambda$ & 1.581  & 140.107  & 18.482   & 24.895   \\
    $Y/E$     & 0.001  & 0.0096   & 0.003052 & 0.002806 \\
    $\nu$     & 0.277  & 0.410    & 0.3322   & 0.0379   \\
    KDF       & 0.100  & 1.089    & 0.562    & 0.207    \\
    \bottomrule
  \end{tabular}
\end{table}

Four commonly used supervised ML algorithms are investigated, namely ANN, SVR, RFR, and HGBoost. These models are selected because they represent complementary learning paradigms capable of capturing complex nonlinear relationships between shell geometry, material properties, and fabrication imperfection quality on the one hand and the KDFs on the other. Additionally, a new hybrid Gaussian Process--XGBoost learning framework (GPR--XGB) is developed to overcome the deficiencies of the aforementioned ML models. Its benefits are demonstrated in terms of prediction accuracy, uncertainty quantification, and extrapolation risk beyond the test dataset. The ML models are described below, and their comparative performance is assessed in terms of the coefficient of determination ($\Rsq$), RMSE, mean absolute error (MAE), and empirical 95\% confidence interval (CI) coverage. Model robustness and the statistical variability of model performance are assessed through repeated 5-fold cross-validation (CV). A total of 50 folds (10 repeats of 5-fold CV) are considered for each training/validation split, and the performance metrics are reported as mean values with the associated error term.

\subsection{Artificial Neural Network (ANN)}
\label{sec:ann}

\begin{figure}[!htbp]
  \centering
  \includegraphics[width=\textwidth]{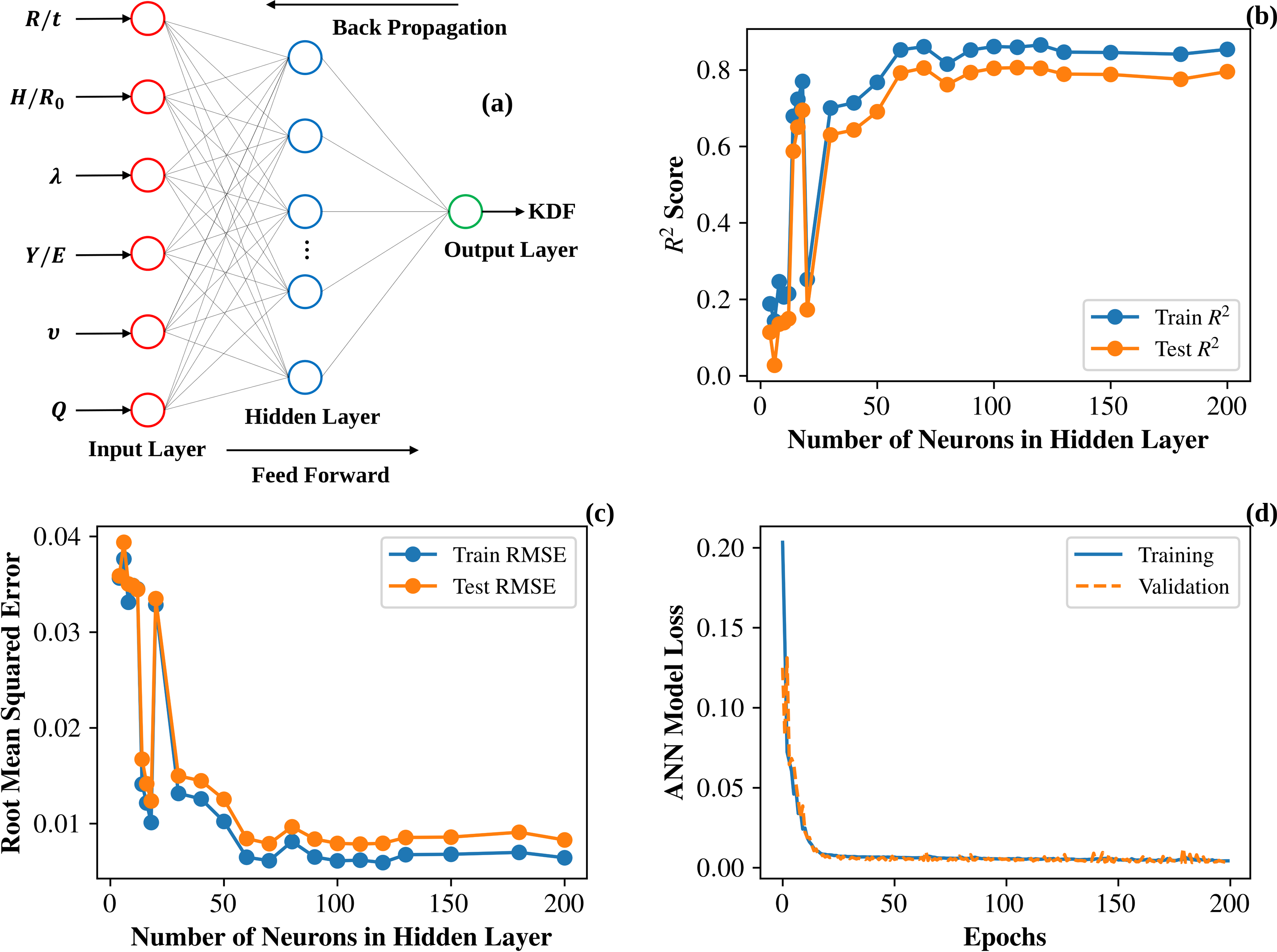}
  \caption{(a) Schematic of the ANN predictive model; sensitivity of (b) $\Rsq$ and (c) RMSE to the number of hidden-layer neurons; and (d) ANN model loss versus number of epochs for the training and validation sets.}
  \label{fig:ann}
\end{figure}

ANNs can serve as powerful predictive modelling tools for estimating the KDFs of thin shells~\citep{majumder2024conical}. In this study, the ANN is used for its ability to learn complex nonlinear relationships between the multiple input parameters affecting the buckling strength reduction of spherical shell caps. The basic structure of an ANN comprises interconnected layers of neurons that iteratively adjust their weights during training to minimize prediction errors on experimental or simulated data, as shown in \cref{fig:ann}(a). The ANN model in this study consists of six input neurons, corresponding to the selected input features, and a single hidden layer. The optimal hidden-layer size (256 neurons) is determined through a sensitivity analysis based on the convergence of $\Rsq$ and RMSE, depicted in \cref{fig:ann}(b)--(c). The Keras API~\citep{gulli2017deep} in Python was used to develop the ANN model. The ReLU activation function~\citep{dubey2019comparative} is used in the hidden layer, whereas a linear activation function is used in the single output layer. The model is trained using the Adam optimizer~\citep{zhang2018improved} with a learning rate of $10^{-3}$, minimizing the RMSE loss. Training is performed for a maximum of 200 epochs with a batch size of 128. To prevent overfitting, early stopping (patience $=50$) is applied based on the validation loss, with automatic restoration of the best-performing model weights. A 10\% validation split is used during training.

\begin{figure}[!htbp]
  \centering
  \includegraphics[width=0.6\textwidth]{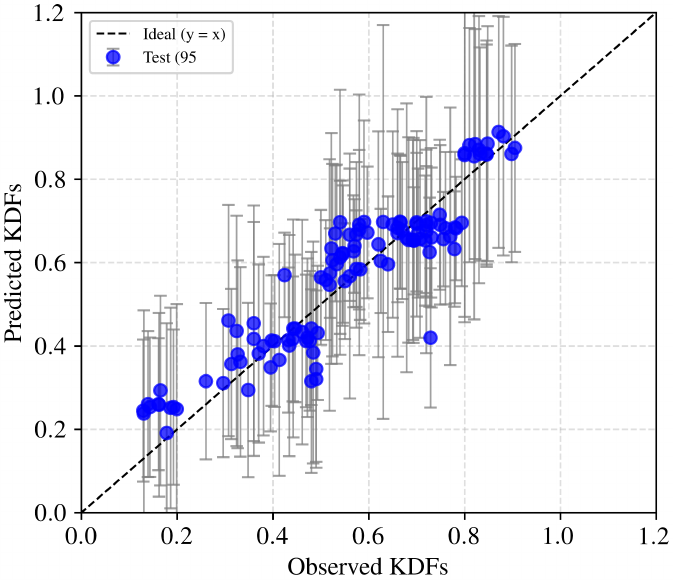}
  \caption{Performance of the ANN model on the test data with 95\% confidence intervals.}
  \label{fig:ann_test}
\end{figure}

\Cref{fig:ann}(d) reports the model losses for the training and validation sets against the epochs. The KDF prediction model yields $\Rsq$, RMSE, and MAE values of $0.86 \pm 0.04$, $0.078 \pm 0.007$, and $0.06 \pm 0.005$, respectively, for the training set, and $\Rsq$ and RMSE values of 0.853 and 0.075, respectively, for the test set.
Moreover, the empirical 95\% CI coverage is 97.8\%.
Thus, the ANN model is fairly accurate in capturing the inherent feature dependencies and their nonlinear relationships with the KDFs, as shown in \cref{fig:ann_test}.

\subsection{Support Vector Regressor (SVR)}
\label{sec:svr}

\begin{figure}[!htbp]
  \centering
  \includegraphics[width=\textwidth]{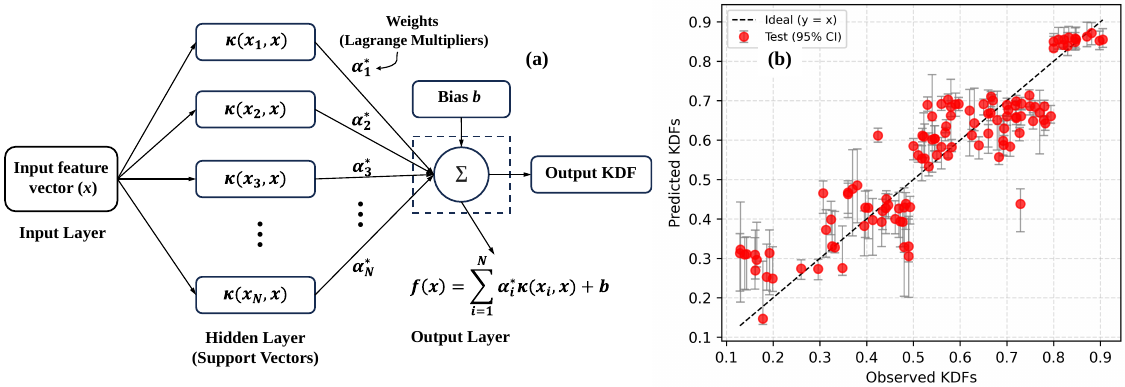}
  \caption{(a) Workflow of the SVR model; (b) performance of the SVR predictive model on the test dataset with 95\% confidence intervals.}
  \label{fig:svr}
\end{figure}

The basic concept of SVR lies in mapping the input variables into a higher-dimensional feature space using kernel functions (e.g., linear, polynomial, or radial basis function (RBF)), followed by establishing an optimal regression hyperplane that best fits the data within a defined margin of tolerance~\citep{zhang2024prediction}. \Cref{fig:svr}(a) shows the generalized workflow of an SVR model, where the input vector $\x$ is mapped into a high-dimensional feature space through kernel evaluations $\kappa(\x_i,\x)$ computed with respect to each support vector $\x_i$. Each kernel output is multiplied by its associated Lagrange multiplier $\alpha_i^{*}$, and the weighted contributions are summed together with the bias term $b$ to produce the final regression output.

The SVR implementation in the Scikit-Learn toolkit was employed in this study. The optimal hyperparameters, determined from a grid search with 5-fold CV, were: (a) an RBF kernel; (b) regularization parameter $C = 20$; (c) $\varepsilon = 0.01$; and (d) $\gamma = $ \texttt{`scale'}. $\Rsq$, RMSE, and MAE values of $0.84 \pm 0.03$, $0.084 \pm 0.006$, and $0.065 \pm 0.005$, respectively, are obtained for the training set, and $\Rsq$ and RMSE values of 0.835 and 0.082 for the test set.
The empirical 95\% CI coverage of 38.4\% signifies very poor calibration of uncertainty in the KDF predictions.
The SVR model fails to account for all sources of variance in the KDF dataset, leading to intervals that are far too narrow. The performance of the SVR model is shown in \cref{fig:svr}(b). The prediction performance of the SVR is found to be inferior to that of the ANN model in all respects.

\subsection{Random Forest Regressor (RFR)}
\label{sec:rfr}

\begin{figure}[!htbp]
  \centering
  \includegraphics[width=\textwidth]{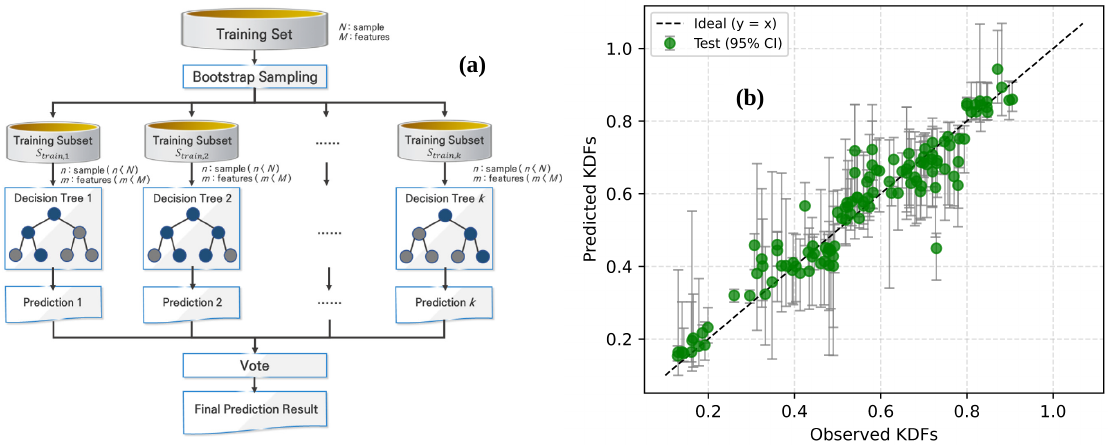}
  \caption{(a) Workflow of the RFR model; (b) performance of the RFR predictive model on the test dataset with 95\% confidence intervals.}
  \label{fig:rfr}
\end{figure}

The RFR is an ensemble ML technique that combines multiple decision trees (DTs) to improve prediction accuracy and robustness~\citep{verma2025prediction}. The basic principle of the RFR involves training numerous DTs on random subsets of the dataset (features and observations) and then averaging their outputs to obtain the final prediction. All DTs are built in parallel with no interaction among them. Inside each DT, the model repeatedly selects the best split (from a random subset of features) that minimizes the prediction error and divides the data into purer subsets. This recursive splitting continues until stopping criteria are met (e.g., maximum depth or minimum number of samples), and each leaf node outputs the average target value of the samples within it, as shown in \cref{fig:rfr}(a). For the present case, the following hyperparameter values were selected based on a repeated 5-fold grid search with CV: (a) a maximum of 200 DTs; (b) no maximum depth criterion for tree growth; and (c) a minimum of 5 samples required to split a node and at least 1 sample per leaf. On the training set, the RFR model predicts the KDFs with $\Rsq$, RMSE, and MAE of $0.898 \pm 0.02$, $0.067 \pm 0.005$, and $0.05 \pm 0.004$, while the test set has $\Rsq$ and RMSE of 0.91 and 0.059, respectively. An empirical 95\% CI coverage of 82.7\% is achieved with this model, indicating decent robustness and calibration against uncertainty in the KDF prediction.
\Cref{fig:rfr}(b) illustrates the performance of the RFR predictive model on the test set with 95\% CIs. The RFR model performs better than the ANN and SVR models in terms of $\Rsq$.

\subsection{Histogram-Based Gradient Boosting (HGBoost)}
\label{sec:hgb}

\begin{figure}[!htbp]
  \centering
  \includegraphics[width=\textwidth]{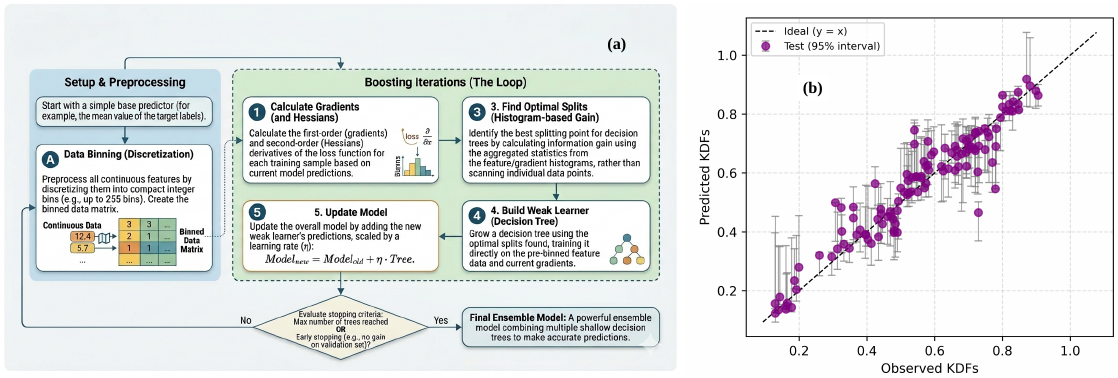}
  \caption{(a) HGBoost model workflow; (b) performance of the HGBoost predictive model on the test data with 95\% confidence intervals.}
  \label{fig:hgb}
\end{figure}

Boosting algorithms build an ensemble of DTs sequentially, each correcting the errors of the prior tree via gradient descent~\citep{duc2023comparison}. This becomes costly for complex data and, unlike the RFR, cannot be parallelized. HGBoost addresses this by binning continuous features into discrete histograms and approximating split points rather than evaluating every feature value, which reduces memory use and training time with little loss of accuracy. This study utilized Scikit-Learn's HGBoost regressor, optimizing the hyperparameters via a repeated 5-fold grid search with CV. The final model used 500 boosting iterations (\texttt{max\_iter} $=500$), a learning rate of 0.2, a maximum tree depth of 10, and a feature fraction of 10\% per split (\texttt{max\_features} $=0.1$) to limit overfitting. The prediction $\Rsq$ and RMSE for the training set are $0.90 \pm 0.018$ and $0.065 \pm 0.004$, respectively, while those for the test set are 0.894 and 0.065, respectively. An empirical prediction coverage of 68.4\% against the nominal 95\% is obtained with the HGBoost model. This signifies moderate-to-poor calibration, with prediction intervals that are too narrow and fail to capture the true underlying uncertainty and data variability. The accuracy of the HGBoost model is better than that of the ANN and SVR models but slightly inferior to that of the RFR model. The model workflow and performance with 95\% CIs are shown in \cref{fig:hgb}(a) and (b), respectively.

\subsection{Hybrid Gaussian Process Regression--XGBoost (GPR--XGB) Model}
\label{sec:gprxgb}

\begin{figure}[!htbp]
  \centering
  \includegraphics[width=0.9\textwidth]{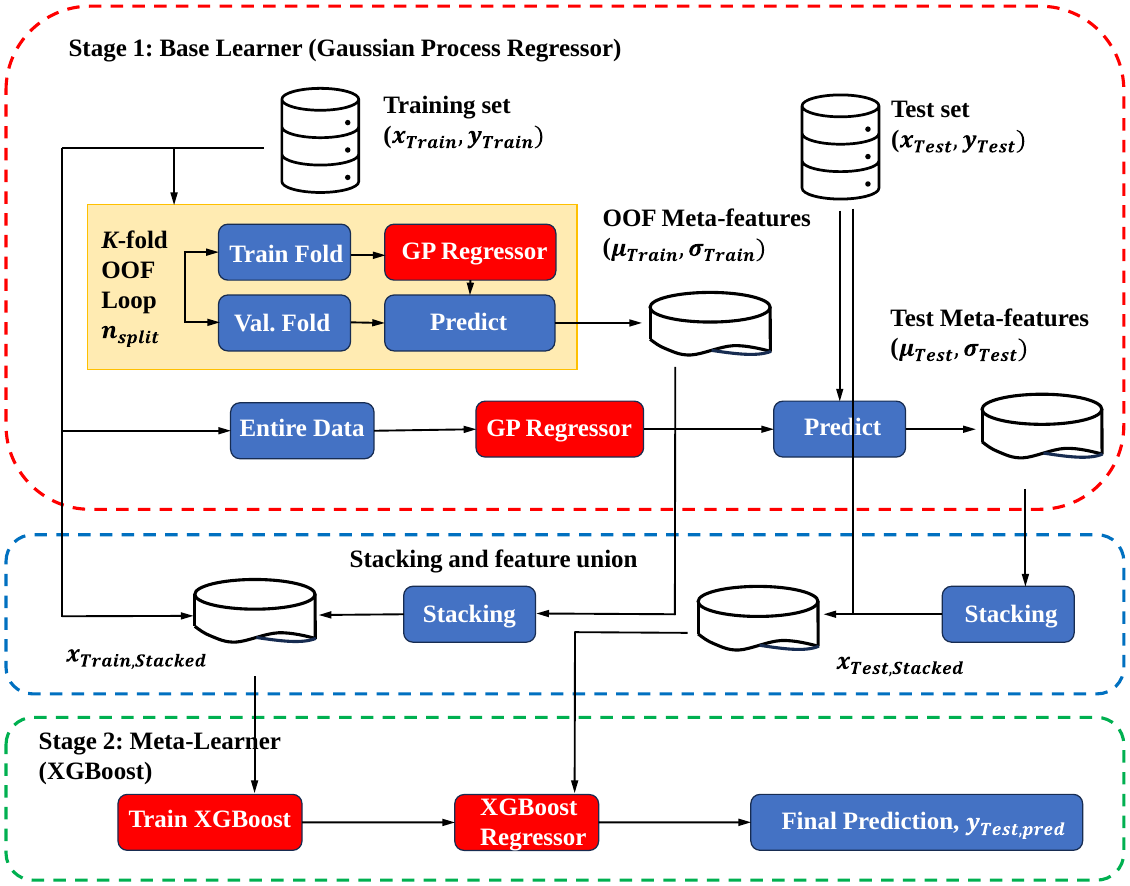}
  \caption{Flow diagram of the hybrid GPR--XGB model for improved KDF prediction.}
  \label{fig:flowchart}
\end{figure}

Gaussian Process Regression (GPR) has been widely acclaimed for probabilistic prediction and uncertainty quantification~\citep{assolie2024advanced}. GPR estimates full predictive distributions, with a mean and dispersion, rather than a single predicted value, and offers flexibility through kernel-based inductive biases that align well with shell buckling behaviour (smooth, but with localized imperfections). However, for complex nonlinear interdependencies among the features and observations, its capabilities can be restricted. The Extreme Gradient Boosting (XGBoost) algorithm is known for its robustness and efficiency in handling highly nonlinear relationships~\citep{ali2023hybrid}; it reduces bias through boosted ensembles and performs well with noisy features. Therefore, in this study, a hybrid two-stage Gaussian Process--XGBoost regressor is explored, combining the complementary strengths of both algorithms to improve the predictive performance for the KDFs. The approach is akin to the authors' previous work on truncated conical shells~\citep{majumder2026data}. In the first stage, the GPR model captures the mean global trends between the KDFs and the input features, along with their uncertainty (dispersion); the GPR (Kriging) metamodel is briefly outlined in \cref{app:gpr} for reference. These results are then used by the XGBoost model in the second stage to learn residual patterns beyond the mean prediction, along with higher-order nonlinear correlations, improving the accuracy of the KDF prediction. The two-stage process can be expressed as
\begin{equation}
  \KDF_{\mathrm{pred}} = f_{\mathrm{GPR}}(\x) + f_{\mathrm{XGB}}\bigl(\x,\mu_{\mathrm{GPR}},\sigma_{\mathrm{GPR}}\bigr) + \varepsilon,
  \label{eq:hybrid}
\end{equation}
where $f_{\mathrm{GPR}}(\x)$ captures the smooth global response surface of the KDFs with respect to the input features $\x$, and, in the second stage, $f_{\mathrm{XGB}}(\x,\mu_{\mathrm{GPR}},\sigma_{\mathrm{GPR}})$ learns the structured residual component that remains unexplained by the kernel model, thereby yielding the final predictor. The GPR--XGB workflow is shown schematically in \cref{fig:flowchart}.

The procedure follows a stacked generalization approach. First, the training dataset is split into training and validation folds, and repeated 5-fold CV is conducted. Within each fold, the GPR is trained on the training split and predicts the mean ($\mu$) and standard deviation ($\sigma$) for the validation split. The results from all 5 folds are then combined to output unbiased meta-features ($\mu_{\mathrm{train}}$, $\sigma_{\mathrm{train}}$) for stacking. This technique is known as out-of-fold (OOF) prediction. Because the OOF meta-feature predictions are obtained from folds that exclude the observation being predicted, they avoid information leakage and are suitable for stacking. For predicting the test-set meta-features ($\mu_{\mathrm{test}}$, $\sigma_{\mathrm{test}}$), however, the entire training set is utilized by the GPR model. There is no risk of data leakage in this step because the test set does not affect model training, and using all the training data yields the best estimates of $\mu_{\mathrm{test}}$ and $\sigma_{\mathrm{test}}$. The Gaussian process kernel is constructed as a composite covariance function,
\begin{equation}
  k(\x,\x') = C\cdot k_{\mathrm{Mat\acute{e}rn},\,\nu=2.5}(\x,\x') + k_{\mathrm{RBF}}(\x,\x') + k_{\mathrm{White}}(\x,\x'),
  \label{eq:kernel}
\end{equation}
where the constant kernel $C$ scales the overall signal variance, the Mat\'ern kernel ($\nu = 2.5$) captures non-smooth realistic variations, and the RBF kernel models smooth global variations~\citep{le2021development}. A white-noise kernel is added to represent experimental noise~\citep{le2021development}. Further details on the kernel composition are given in \cref{app:gpr}.

The mean $\mu_{\mathrm{train}}$ is used to construct leakage-free residual targets $r_{\mathrm{train}} = \KDF_{\mathrm{train}} - \mu_{\mathrm{train}}$, and the meta-features ($\mu_{\mathrm{train}}$, $\sigma_{\mathrm{train}}$) are appended to the training feature set to generate an uncertainty-informed stacked design matrix for the XGBoost residual learner. The second-stage XGBoost regressor represents an additive ensemble of $K$ regression trees mapping the stacked feature vector $\z_i = [\x_i, \mu_i, \sigma_i]$---the original input features augmented with the GPR-derived mean and standard-deviation meta-features---to the residual target $r_i$. The hyperparameters of the XGBoost learner are selected by a repeated randomized cross-validated search (using $\X_{\mathrm{train,stacked}}$ and $r_{\mathrm{train}}$), yielding the following optimal values: (a) 1000 estimators; (b) a learning rate of 0.01; (c) no limit on the maximum depth; (d) a subsample proportion of 0.8; (e) a random fraction of 80\% of the features used to construct each tree; and (f) L1 and L2 regularization constants of 1.0. For the test set, a final GPR is fit on the entire training set to predict the mean and standard deviation for the test set, which are then stacked with the test features to obtain the stacked test features:
\begin{subequations}
\label{eq:stack}
\begin{align}
  \X_{\mathrm{test,stacked}} &= \bigl[\X_{\mathrm{test}},\, \mu_{\mathrm{test}},\, \sigma_{\mathrm{test}}\bigr]. \label{eq:stack_a}
\intertext{The tuned XGBoost model is then used to predict the residuals for the test set,}
  \check{r}_{\mathrm{test}} &= f_{\mathrm{XGB}}\bigl(\X_{\mathrm{test,stacked}}\bigr), \label{eq:stack_b}
\intertext{and the final hybrid KDF prediction is obtained as}
  \widehat{\KDF}_{\mathrm{test}} &= \mu_{\mathrm{test}} + \check{r}_{\mathrm{test}}. \label{eq:stack_c}
\end{align}
\end{subequations}

To quantify predictive uncertainty on unseen data, split-conformal prediction intervals are constructed using the absolute out-of-fold hybrid residuals, with a finite-sample calibrated quantile correction to produce nominal 95\% intervals.

\begin{figure}[!htbp]
  \centering
  \includegraphics[width=0.65\textwidth]{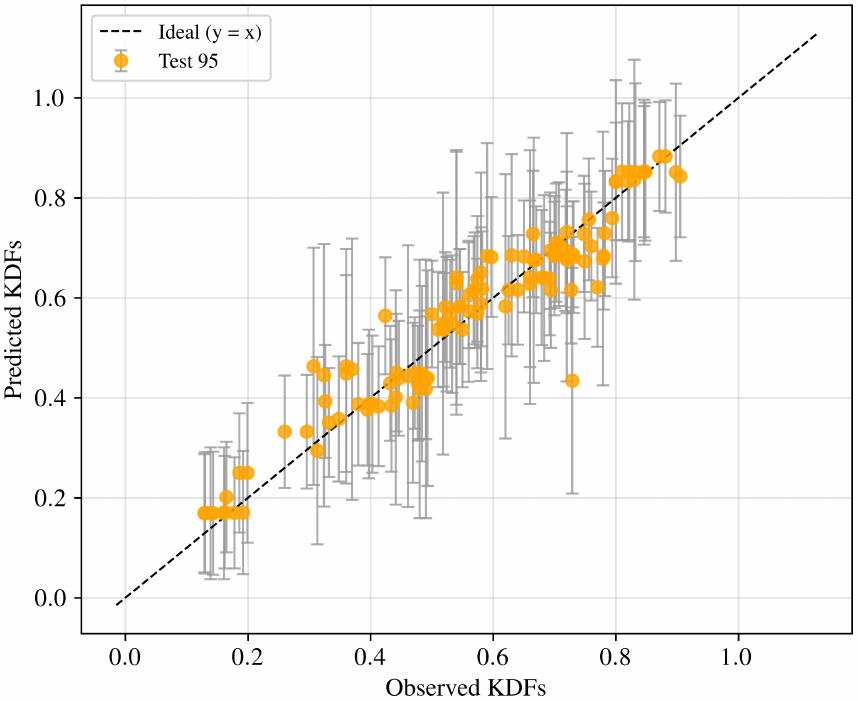}
  \caption{Prediction performance of the hybrid GPR--XGB model on the test dataset with 95\% confidence intervals.}
  \label{fig:gprxgb_test}
\end{figure}

The XGBoost model captures residual patterns and benefits from the additional uncertainty information provided by the GPR stage. The hybrid model performs best among all ML models, obtaining an $\Rsq$ of $0.91 \pm 0.015$, RMSE of $0.063 \pm 0.006$, and MAE of $0.046 \pm 0.004$ on the training set. On the test set, the results are similarly superior to those of all other models, with $\Rsq$ and RMSE of 0.93 and 0.056, respectively. Additionally, the empirical 95\% CI coverage is 96.7\%, which is higher than that of all models except the ANN. \Cref{fig:gprxgb_test} shows the prediction results on the test set with 95\% CIs.

Thus, in terms of KDF prediction accuracy and quantification of the associated uncertainty, the hybrid model performs as well as or better than the other popular ML models considered. This model therefore shows potential as a useful data-driven tool for assessing the buckling behaviour of thin shells.

\section{Performance Comparison of the ML Models in Predicting the KDFs}
\label{sec:comparison}

The KDF predictive performance of the ANN, SVR, RFR, HGBoost, and GPR--XGB models on the held-out test set is compared in \cref{tab:performance} in terms of $\Rsq$, RMSE, MAE, and 95\% CI coverage. Additionally, to assess the performance of simpler models for KDF prediction, comparisons with a Kriging-only model (i.e., GPR alone) and a conventional polynomial response surface metamodel (RSM) are also made; a brief overview of these two models is provided in \cref{app:gpr,app:rsm}. Among all models, the GPR--XGB hybrid model achieves the best overall predictive accuracy, with the highest $\Rsq$ of 0.93, the lowest RMSE of 0.056, and the lowest MAE of 0.042. Compared with the ANN, GPR--XGB improves $\Rsq$ from 0.853 to 0.93, an increase of approximately 9\%, while reducing RMSE from 0.078 to 0.056 and MAE from 0.062 to 0.042, corresponding to reductions of approximately 28.2\% and 32.3\%, respectively. Relative to the SVR, which shows the weakest performance ($\Rsq = 0.81$, RMSE $=0.087$, MAE $=0.069$), the GPR--XGB model increases $\Rsq$ by approximately 14.8\% and reduces RMSE and MAE by approximately 35.6\% and 39.1\%, respectively. The RFR also shows strong performance ($\Rsq = 0.906$, RMSE $=0.062$, MAE $=0.044$); however, GPR--XGB still reduces RMSE by approximately 9.7\% and MAE by approximately 4.5\% relative to the RFR. Similarly, GPR--XGB outperforms HGBoost, increasing $\Rsq$ from 0.893 to 0.93 and reducing RMSE from 0.065 to 0.056. In terms of uncertainty representation, the ANN and GPR--XGB achieve the highest 95\% CI coverage values of 97.4\% and 96.7\%, respectively, indicating reliable uncertainty bounds. Although the ANN provides slightly higher coverage, its prediction errors are substantially larger than those of GPR--XGB. The SVR shows a poor CI coverage of only 30.8\%, suggesting unreliable uncertainty estimation. The Kriging-only model also demonstrates competitive performance, achieving $\Rsq = 0.89$, RMSE $=0.062$, MAE $=0.045$, and a 95\% CI coverage of 95.2\%. Compared with Kriging-only, GPR--XGB increases $\Rsq$ by approximately 4.5\% while reducing RMSE and MAE by approximately 9.7\% and 6.7\%, respectively. The polynomial RSM shows relatively weaker predictive accuracy, with $\Rsq = 0.817$, RMSE $=0.085$, and MAE $=0.069$.
Relative to the polynomial RSM, GPR--XGB increases $\Rsq$ by approximately 13.8\% and reduces RMSE and MAE by approximately 34.1\% and 39.1\%, respectively. Overall, the GPR--XGB model provides the best balance between predictive accuracy and uncertainty coverage, making it the most reliable of the models evaluated.

\begin{table}[!htbp]
  \centering
  \caption{Performance of the ML predictive models on the held-out test set.}
  \label{tab:performance}
  \begin{tabular}{lcccc}
    \toprule
    Model & $\Rsq$ & RMSE & MAE & 95\% CI coverage (\%) \\
    \midrule
    ANN            & 0.853 & 0.078 & 0.062 & 97.5 \\
    SVR            & 0.810 & 0.087 & 0.069 & 30.8 \\
    RFR            & 0.906 & 0.062 & 0.044 & 82.1 \\
    HGBoost        & 0.893 & 0.065 & 0.044 & 68.4 \\
    \textbf{GPR--XGB} & \textbf{0.930} & \textbf{0.056} & \textbf{0.042} & 96.7 \\
    Kriging-only   & 0.890 & 0.062 & 0.045 & 95.2 \\
    Polynomial RSM & 0.817 & 0.095 & 0.069 & --- \\
    \bottomrule
  \end{tabular}
\end{table}

\begin{figure}[!htbp]
  \centering
  \includegraphics[width=\textwidth]{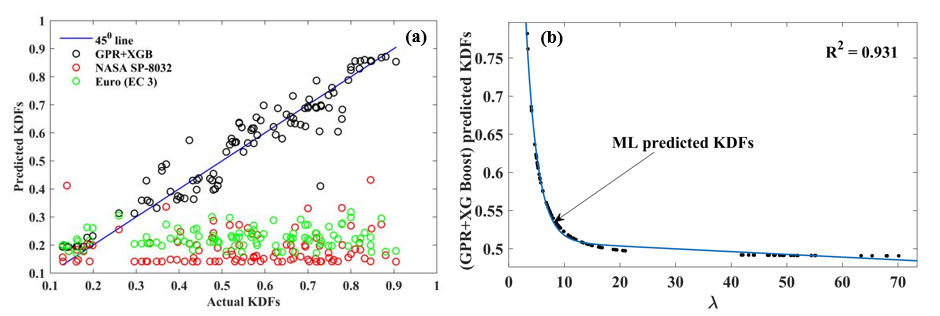}
  \caption{(a) GPR--XGB-predicted and code-stipulated KDFs versus actual KDFs; (b) GPR--XGB-predicted KDFs versus $\lambda$.}
  \label{fig:kdf_codes}
\end{figure}

The KDFs predicted by the GPR--XGB model and those stipulated by design codes (NASA SP-8032~\citep{nasa1969sp8032} and Eurocode~3 (EC3)~\citep{rotter2013buckling}) are plotted against the actual (experimentally observed) KDFs in \cref{fig:kdf_codes}(a). In contrast to the accurate and robust KDF predictions of the GPR--XGB model, the NASA SP-8032~\citep{nasa1969sp8032} and EC3~\citep{rotter2013buckling} predictions remain substantially lower (20--40\% on the conservative side) than the actual values for most data points, highlighting their conservative nature. The variation of the ML-predicted KDFs with $\lambda$ is shown in \cref{fig:kdf_codes}(b). The ML-predicted KDFs lie within $\pm5\%$ of the actual KDFs and may be effectively employed within the limits of the parametric range. The ML-predicted KDFs exhibit a clear nonlinear decreasing trend with increasing shell shape parameter $\lambda$, with a rapid reduction at lower $\lambda$ followed by a gradual plateau at higher $\lambda$. Thus, the proposed GPR--XGB model captures the dominant dependence of the KDF on $\lambda$ with good accuracy. Overall, the results suggest that the hybrid model provides a less conservative yet accurate alternative to traditional code-based formulations for predicting shell buckling KDFs.

\subsection{Calibration of KDF Prediction Uncertainty and Confidence}
\label{sec:calibration}

\begin{figure}[!htbp]
  \centering
  \includegraphics[width=0.65\textwidth]{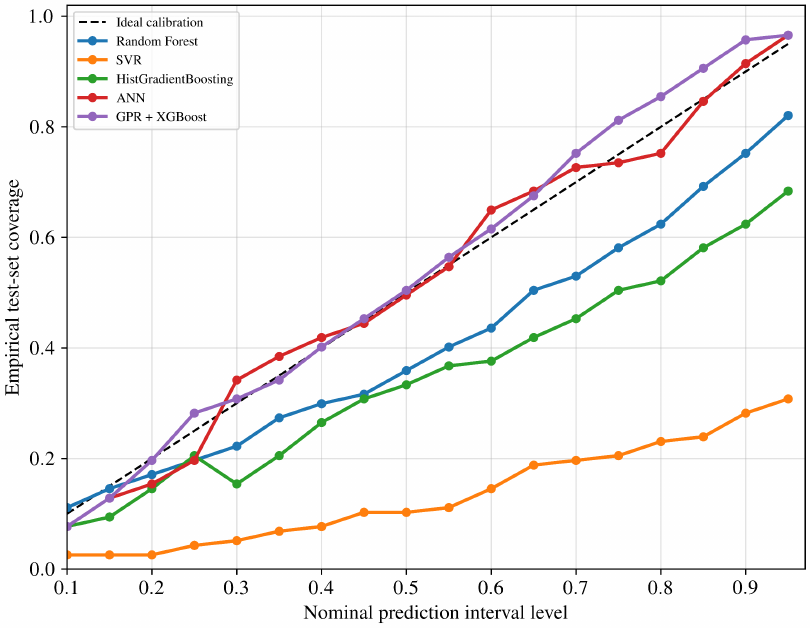}
  \caption{Calibration curves showing nominal versus empirical confidence interval coverage (10\%--95\%) for the KDFs predicted by all ML models.}
  \label{fig:calibration}
\end{figure}

As mentioned in the individual model descriptions, uncertainty quantification is performed for all models to evaluate the reliability of the predicted KDFs. Prediction intervals are generated for each model and assessed using empirical coverage over the test set. This section presents a more in-depth assessment of the KDF prediction uncertainty and confidence. Calibration curves are constructed by comparing the nominal prediction interval level with the corresponding empirical test-set coverage for interval levels ranging from 10\% to 95\%, as shown in \cref{fig:calibration}.

\Cref{fig:calibration} shows that the proposed GPR--XGB model offers the best uncertainty estimates among the models considered. Its empirical coverage closely follows the ideal calibration line over the full range of nominal prediction interval levels and becomes only mildly conservative at high confidence levels. At the 95\% nominal level, the GPR--XGB model achieves near-complete empirical coverage, whereas the RFR, HGBoost, and SVR show clear under-coverage, with empirical coverages of 0.82, 0.68, and 0.30, respectively. The ANN curve closely follows the ideal calibration up to 70\%, but in the nominal range of 75\%--85\% its coverage drops below the ideal requirement. It then increases and modestly outperforms the GPR--XGB curve at 95\% nominal coverage (97.53\% versus 96.7\%). Overall, the calibration analysis indicates that the GPR--XGB model provides the best balance of reliability and conservatism. It avoids the strong under-coverage observed for the SVR, HGBoost, and RFR models, while maintaining near-nominal to mildly conservative coverage at high confidence levels. This supports the use of the GPR--XGB model for uncertainty-aware and reliability-oriented KDF prediction.

\begin{figure}[!htbp]
  \centering
  \includegraphics[width=\textwidth]{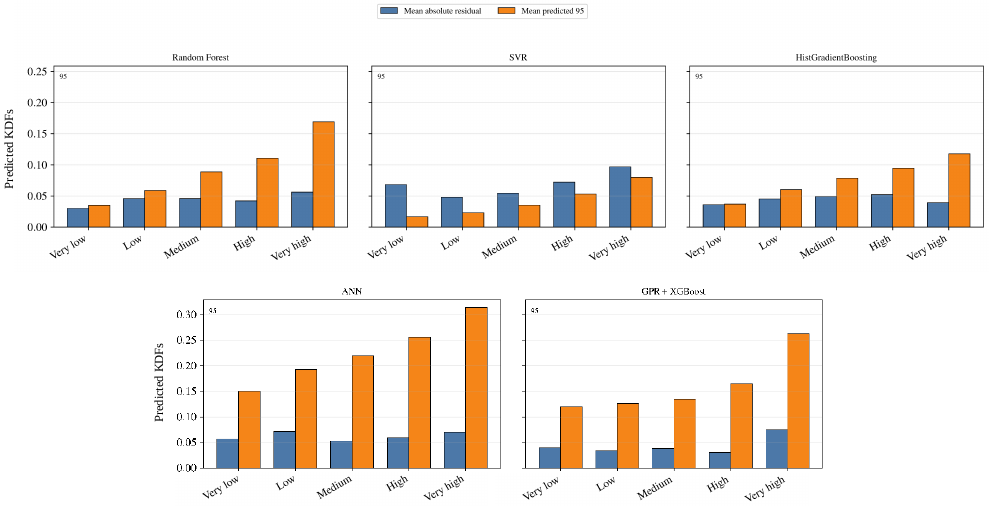}
  \caption{Mean absolute KDF residuals versus ML-predicted 95\% confidence half-widths in different uncertainty bands.}
  \label{fig:bands}
\end{figure}

To further assess whether the predicted KDF uncertainty is informative at the sample level, the test samples are divided into five uncertainty bands: very low, low, medium, high, and very high. Within each band, the mean absolute KDF residual is compared with the mean predicted 95\% half-width (\cref{fig:bands}). For the SVR, the predicted 95\% half-widths are frequently smaller than the observed mean absolute residuals, which is consistent with its poor empirical coverage and indicates overconfident uncertainty estimates. The RFR and HGBoost show increasing interval widths across the uncertainty bands, but their calibration curves indicate that these intervals remain insufficiently wide to achieve nominal coverage. The ANN provides strong coverage, but its predicted 95\% half-widths are comparatively large, increasing from approximately 0.15 in the very-low uncertainty band to more than 0.30 in the very-high band, indicating reduced sharpness. The GPR--XGB model provides a more balanced KDF uncertainty estimate: its predicted 95\% half-widths increase systematically from approximately 0.12 to 0.26 from the very-low to the very-high uncertainty bands, while the corresponding mean absolute residuals remain lower, at approximately 0.03--0.08. The very-high uncertainty band also contains the largest residuals, indicating that the model assigns greater uncertainty to the KDF samples that are genuinely more difficult to predict. Overall, GPR--XGB avoids the under-coverage observed for the SVR, RFR, and HGBoost, while providing sharper intervals than the ANN. This demonstrates that the proposed hybrid model offers calibrated, slightly conservative, but practically informative uncertainty estimates for reliability-oriented KDF prediction.

\subsection{Dataset Sparsity and Extrapolation Risk beyond the Observed Feature Space}
\label{sec:sparsity}

The ML models were trained on a finite experimental database; hence, the reliability of their KDF predictions depends on the distribution and density of the available data in the input-feature space. To assess the applicability domain of the GPR--XGB model, the input variables were standardized, and local data sparsity was quantified using the nearest-neighbour distance in the standardized feature space. Each test sample was assigned to one of five sparsity quintiles based on its mean distance to the nearest training samples: Q1 corresponds to the densest regions of the training database, whereas Q5 corresponds to the sparsest.

\begin{figure}[!htbp]
  \centering
  \begin{tikzpicture}
    \node[anchor=south west,inner sep=0] (img) {\includegraphics[width=\textwidth]{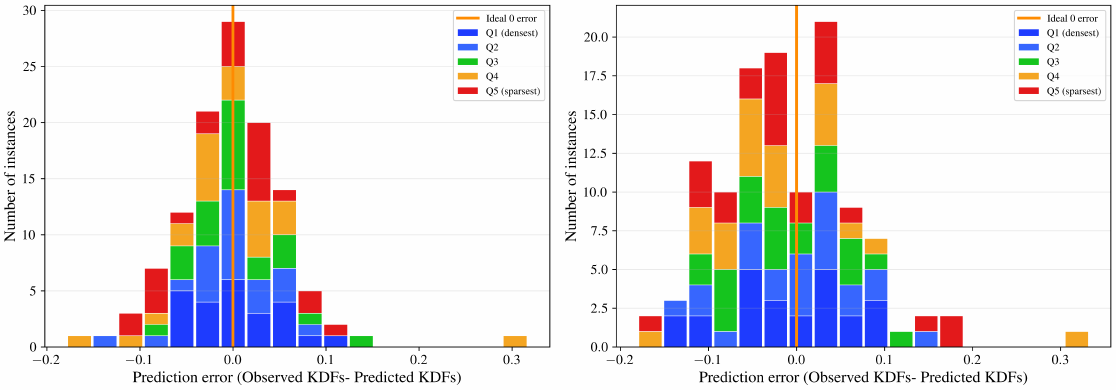}};
    \begin{scope}[x={(img.south east)},y={(img.north west)}]
      \node[font=\bfseries\small] at (0.115,0.90) {(a)};
      \node[font=\bfseries\small] at (0.605,0.90) {(b)};
    \end{scope}
  \end{tikzpicture}
  \caption{KDF prediction-error distribution stratified by local data sparsity for (a) the GPR--XGB model and (b) the RFR model.}
  \label{fig:sparsity}
\end{figure}

The KDF prediction-error distribution stratified by sparsity quintile is shown in \cref{fig:sparsity} for the GPR--XGB and RFR models. The prediction error is defined as the difference between the observed and model-predicted KDF, and the vertical reference line represents zero error. For the GPR--XGB model, the distribution is centred near zero, indicating limited systematic bias in the model predictions. Most samples, including those from relatively sparse regions, are concentrated close to the zero-error line, suggesting that the model performs reliably within the experimentally represented parameter space. The red Q5 segments, representing the sparsest regions, are mostly located near the central error range, roughly between $-0.1$ and $0.05$. However, a small number of larger errors appear in the distribution tails, including an isolated positive-error point near 0.30 and some negative-error cases near $-0.15$ to $-0.20$, indicating increased uncertainty in sparsely sampled or boundary regions.

For comparison, the RFR also shows an error distribution centred broadly around zero, but the distribution is wider and more asymmetric than that of the GPR--XGB model. For the RFR model, a larger fraction of the errors is concentrated in the negative range, approximately between $-0.15$ and $-0.05$, indicating more frequent overprediction of the KDF. The RFR distribution also shows more visible sparse-region contributions in the tails, particularly from Q4 and Q5 samples. In contrast, the GPR--XGB model has its dominant error mass concentrated within approximately $\pm0.08$ of zero, and the sparsest Q5 samples remain mostly embedded in the central error region. This indicates that, although both models can produce occasional tail errors, the GPR--XGB model exhibits a more compact and better-centred error distribution with lower sensitivity to local sparsity.

Therefore, the trained GPR--XGB model should be considered most reliable for interpolation within the range of parameter combinations represented in the experimental database. Predictions outside this applicability domain should be interpreted cautiously and accompanied by the calibrated prediction intervals. The comparison with the RFR further supports this interpretation: while the RFR provides competitive point-prediction performance, the GPR--XGB model provides a more favourable error structure in sparse regions, suggesting a lower extrapolation risk within the experimentally represented feature space.

\subsection{Global Feature Importance Using SHapley Additive exPlanations (SHAP)}
\label{sec:shap}

\begin{figure}[!htbp]
  \centering
  \includegraphics[width=\textwidth]{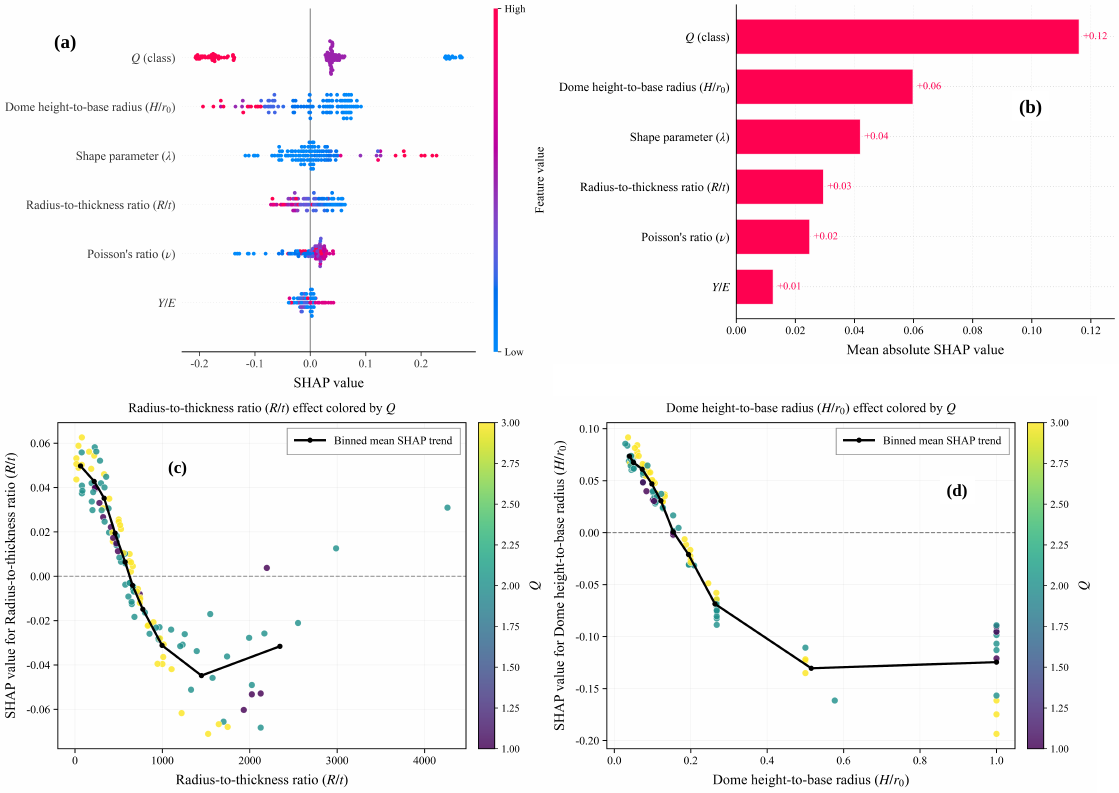}
  \caption{Sensitivity analysis of the hybrid GPR--XGB model: (a) SHAP bee-swarm plot; (b) mean absolute SHAP values; and SHAP dependence of (c) $R/t$ and (d) $H/r_0$, coloured by fabrication quality $Q$.}
  \label{fig:shap}
\end{figure}

To assess the relative importance of the different features for KDF prediction, a SHAP (SHapley Additive exPlanations) analysis, introduced by \citet{lundberg2017unified}, is carried out; efficient algorithms for estimating the underlying Shapley value feature attributions are reviewed by \citet{chen2023algorithms}.
SHAP is particularly suitable for the present study because the KDF depends on coupled geometric, material, and fabrication-related variables. The SHAP analysis is conducted at two complementary levels: global feature influence is first evaluated using a bee-swarm plot and then a mean-absolute SHAP bar chart. \Cref{fig:shap}(a) presents the SHAP bee-swarm summary plot for the optimized GPR--XGB model, while \cref{fig:shap}(b) shows the corresponding global feature ranking based on the mean absolute SHAP values. Together, these visualizations provide both local and global interpretations of the model predictions. The bee-swarm plot illustrates the magnitude, direction, and distribution of the SHAP contribution of each observation, whereas the mean absolute SHAP values quantify the average influence of each input variable on the predicted KDF. The SHAP analysis identifies the fabrication tolerance quality parameter $Q$ as the most influential predictor, with a mean absolute SHAP value of approximately 0.12, followed by the dome height-to-base radius ratio $H/r_0$, the shell shape parameter $\lambda$, the radius-to-thickness ratio $R/t$, Poisson's ratio $\nu$, and the normalized yield-strength ratio $Y/E$.

The resulting ranking is physically consistent with the mechanics of imperfection-sensitive shell buckling. Fabrication quality dominates because it serves as an indirect measure of the severity of the geometric imperfections that govern knockdown behaviour. Among the geometric variables, $H/r_0$, $\lambda$, and $R/t$ describe the shell geometry and slenderness, which strongly influence buckling susceptibility, whereas the comparatively smaller SHAP values associated with $\nu$ and $Y/E$ indicate that material properties play a secondary role within the investigated experimental domain.

\subsection{SHAP Dependence Analysis of Geometry--Fabrication Interactions}
\label{sec:shap_dep}

While the global SHAP analysis establishes the relative importance of the input variables, it does not reveal how their contributions vary across the design space or with fabrication quality. To investigate these aspects, SHAP dependence plots were generated for $R/t$ and $H/r_0$, as shown in \cref{fig:shap}(c) and (d), respectively. In these plots, the abscissa represents the geometric variable, the ordinate denotes its SHAP contribution to the GPR--XGB prediction relative to the model baseline, and the marker colour indicates the fabrication-quality class $Q$. The solid black curve represents the binned mean SHAP trend. Positive SHAP values indicate that the corresponding geometric variable increases the predicted KDF relative to the baseline, whereas negative values indicate a reduction. Consequently, the dependence plots describe the conditional contribution learned by the GPR--XGB model rather than the direct functional relationship between the KDF and an individual geometric parameter.

\Cref{fig:shap}(d) shows that the SHAP contribution associated with $H/r_0$ decreases nonlinearly over the investigated range. For $H/r_0 \le 0.10$, the SHAP contribution is predominantly positive (approximately 0.03--0.09); it crosses zero near $H/r_0 \approx 0.17$, decreases to approximately $-0.07$ around $H/r_0 \approx 0.27$, and gradually approaches a nearly constant value close to $-0.13$ at larger $H/r_0$. This evolution is consistent with the mechanics of imperfection-sensitive shell buckling. Increasing $H/r_0$ generally increases the shell's susceptibility to instability, resulting in lower KDFs. The reduced slope observed at larger $H/r_0$ suggests that, once the geometric contribution has shifted substantially below the model baseline, additional increases in $H/r_0$ produce comparatively smaller incremental changes in the predicted KDF.

A similar trend is observed for $R/t$ in \cref{fig:shap}(c). For relatively stocky shells ($R/t \le 600$), the SHAP contribution remains positive (approximately 0.02--0.06); it crosses zero near $R/t \approx 650$, reaches a minimum of approximately $-0.045$ around $R/t \approx 1500$--1700, and then increases slightly toward approximately $-0.03$ while remaining negative. This behaviour is also mechanically consistent with classical thin-shell theory, in which increasing $R/t$ reduces the relative bending stiffness and increases sensitivity to unavoidable geometric imperfections, thereby lowering the attainable KDF. The slight flattening of the SHAP trend at larger $R/t$ indicates that, once the shell has entered a highly imperfection-sensitive regime, additional increases in slenderness modify the predicted contribution less rapidly than during the transition from relatively thick to thin shells.

The colour distribution provides additional insight into geometry--fabrication interactions. At comparable values of $H/r_0$ and $R/t$, specimens belonging to different fabrication-quality classes occupy different vertical SHAP positions. For example, within $R/t \approx 700$--1000, the SHAP contributions vary from approximately $-0.01$ to $-0.03$ across the fabrication classes, with comparable dispersion observed over intermediate $H/r_0$ values. This indicates that fabrication quality modulates the geometric contribution rather than acting as an independent additive correction, reflecting the role of manufacturing imperfections in governing shell buckling. Nevertheless, the fabrication-quality classes remain substantially intermingled rather than forming distinct colour-separated bands, indicating that the learned relationship is governed primarily by nonlinear geometric effects, with fabrication quality providing a secondary modulation. Accordingly, the dependence plots provide descriptive evidence of geometry--fabrication interactions without implying exact pairwise SHAP interaction values or causal relationships. Overall, the GPR--XGB model captures physically meaningful nonlinear geometric behaviour while accounting for specimen-to-specimen variability associated with fabrication quality.

\section{Reliability-Based Design for Safety-Consistent KDFs of Thin Spherical Shell Caps}
\label{sec:rbd}

For a spherical shell cap subjected to external pressure, the design must ensure that the critical buckling pressure exceeds the applied pressure demand. The influence of the governing shell parameters, such as $R/t$, $H/r_0$, $\lambda$, and the fabrication quality class $Q$, on the KDF has been demonstrated in the preceding sections using different ML predictive models. However, owing to the inherent uncertainties associated with the geometric parameters, fabrication surface imperfections, and model prediction, it is more appropriate to incorporate reliability-based measures in KDF selection. Accordingly, the design KDF can be calibrated to satisfy a prescribed target reliability level through an RBD framework.

In conventional reliability analysis and RBD procedures, the KDF must be evaluated repeatedly at each trial design point during the numerical optimization used to estimate the reliability index $\beta$. This process can become computationally expensive. To overcome this limitation, the ML surrogate is employed as a metamodel to efficiently establish a functional mapping between the input parameters and the resulting KDFs. In the present study, two ML-based metamodels were developed (one with GPR--XGB and the other with the ANN) for a varying shell shape parameter $\lambda$, which is also the chosen design variable (DV), within the range $5 \le \lambda \le 40$, and fabrication quality class $Q$.

\subsection{Development of the ML-Based Metamodel}
\label{sec:metamodel}

\begin{figure}[!htbp]
  \centering
  \includegraphics[width=0.75\textwidth]{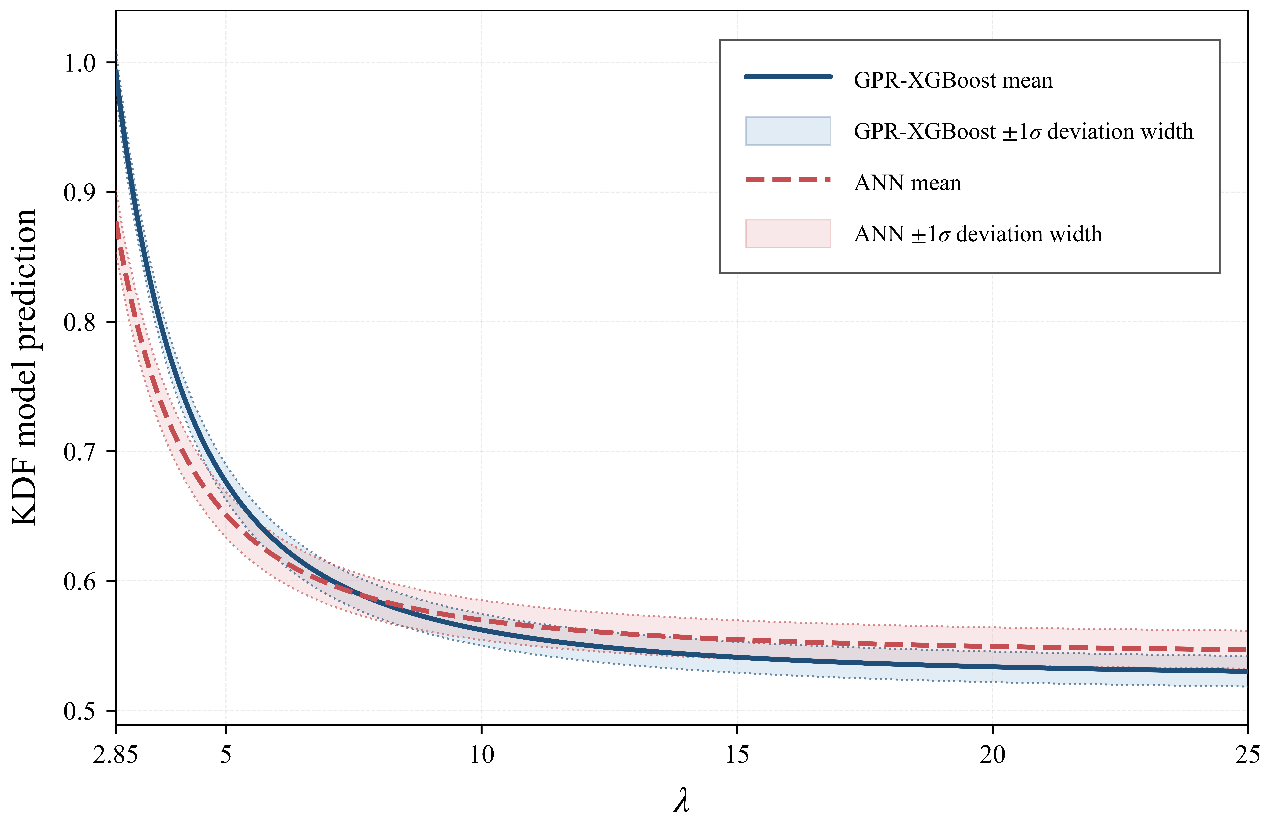}
  \caption{Variation of the predicted KDF with $\lambda$ for the GPR--XGB hybrid and ANN metamodels, showing the mean ($\mu$) prediction and the corresponding $\pm1\sigma$ uncertainty bounds.}
  \label{fig:metamodel}
\end{figure}

Since the test-set KDF predictions are associated with 95\% prediction intervals, for each investigated value of the shell shape parameter $\lambda$, 10{,}000 independent predictions are generated by fixing $\lambda$ while independently resampling the remaining input variables from their empirical distributions obtained from the test database. This gives a conditional prediction distribution that captures the variability in the predicted KDF arising from the combined geometric, material, and fabrication-related uncertainties represented in the data. At each $\lambda$, the Monte Carlo predictions are summarized by their sample mean together with the corresponding one-standard-deviation ($\pm1\sigma$) limits, and continuous analytical response functions are constructed using nonlinear least-squares regression. Following NASA SP-8032~\citep{nasa1969sp8032}, the Monte Carlo mean response is approximated using the inverse-square relationship $\KDF(\lambda) = a + b/\lambda^{2}$. The same fitting procedure is performed independently for the upper ($\mu + \sigma$) and lower ($\mu - \sigma$) uncertainty envelopes, resulting in three continuous response functions for each surrogate model. \Cref{fig:metamodel} compares the analytical response curves obtained for the proposed GPR--XGB and ANN metamodels.

The corresponding regression equations, with coefficients $a$ and $b$, together with the mean uncertainty bandwidth, are summarized in \cref{tab:fits}.

\begin{table}[!htbp]
  \centering
  \caption{Fitted inverse-square response functions and corresponding mean prediction uncertainty bandwidths ($\pm1\sigma$) for the ANN and GPR--XGB metamodels.}
  \label{tab:fits}
  \renewcommand{\arraystretch}{1.2}
  \begin{tabular}{lcccc}
    \toprule
    Model & Mean ($\mu$) & $\mu + \sigma$ & $\mu - \sigma$ & Mean bandwidth ($\pm1\sigma$) \\
    \midrule
    ANN      & $0.543 + 2.714/\lambda^{2}$ & $0.557 + 2.790/\lambda^{2}$ & $0.528 + 2.636/\lambda^{2}$ & 0.30 \\
    GPR--XGB & $0.524 + 3.800/\lambda^{2}$ & $0.536 + 3.846/\lambda^{2}$ & $0.512 + 3.754/\lambda^{2}$ & 0.23 \\
    \bottomrule
  \end{tabular}
\end{table}

Both metamodels recover the expected reduction in KDF with increasing $\lambda$, characterized by a rapid decrease at relatively small values of $\lambda$ followed by a gradual transition toward an asymptotic response at larger values. This trend is consistent with the classical mechanics of imperfection-sensitive shell buckling, where increasing shell slenderness progressively reduces the attainable buckling strength. \Cref{fig:metamodel} shows that the GPR--XGB metamodel exhibits a consistently narrower prediction uncertainty than the ANN metamodel, particularly in the low-$\lambda$ region ($\lambda \approx 2.85$--8), where the KDF is most sensitive to changes in geometry. Beyond $\lambda \approx 8$, the ANN mean response remains consistently above the GPR--XGB prediction, while both models approach nearly constant uncertainty bands; however, the GPR--XGB maintains a smaller $\pm1\sigma$ bandwidth throughout the investigated range, particularly in the highly sensitive region, reinforcing that the hybrid model produces more stable KDF predictions with reduced predictive uncertainty.

\subsection{RBD and Safety-Consistent KDFs through the First-Order Reliability Method (FORM)}
\label{sec:form}

The ML-based metamodels are subsequently used for repeated evaluations of the limit state function (LSF), expressed as
\begin{equation}
  g\bigl(\KDF(\lambda),\KDFd(\lambda)\bigr) = \KDF(\lambda) - \KDFd(\lambda),
  \label{eq:lsf}
\end{equation}
where $\KDF$ is the actual ML-predicted KDF, expressed as a function of $\lambda$ ($5 \le \lambda \le 35$) and treated as a random variable, while $\KDFd$ is the design KDF recommended by codal provisions (NASA SP-8032~\citep{nasa1969sp8032}), as given in \cref{eq:kdf_a,eq:kdf_b}, respectively:
\begin{subequations}
\label{eq:kdf}
\begin{align}
  \KDF  &= 1.006\,\lambda^{-1.294} + 0.4612, \label{eq:kdf_a}\\
  \KDFd &= 0.14 + \frac{3.2}{\lambda^{2}}. \label{eq:kdf_b}
\end{align}
\end{subequations}

The failure surface is defined by the LSF that separates the safe and unsafe domains:
\begin{equation}
  g\bigl(\KDF(\lambda),\KDFd(\lambda)\bigr) = 0.
  \label{eq:failure_surface}
\end{equation}
The safe domain corresponds to $g > 0$, whereas failure occurs when $g < 0$. Thus, failure corresponds to the condition in which the ML-predicted KDF is smaller than the design KDF required by the selected code provision. The reliability index $\beta$ represents the shortest distance between the origin of the reduced coordinate system and the limit-state surface in standard normal space; the corresponding point on the limit-state surface is referred to as the design point. For nonlinear or complex LSFs, determining $\beta$ and locating the design point requires solving a constrained optimization problem, which can be written as
\begin{subequations}
\label{eq:form}
\begin{align}
  \beta &= \min_{\{\chi\}} \sqrt{\{\chi\}^{\T}\{\chi\}}, \label{eq:form_a}\\
  \text{subject to}\quad & g\bigl(\KDF(\lambda),\KDFd(\lambda)\bigr) = 0, \label{eq:form_b}
\end{align}
\end{subequations}
where $\{\chi\}$ denotes the vector of transformed random variables in standard normal space. The Rackwitz--Fiessler algorithm is commonly used to solve this optimization problem through iterative linearization of the LSF~\citep{haldar2000probability}. In the present study, the optimization was performed using the pattern search algorithm in MATLAB R2024a. Once $\beta$ is obtained, the probability of failure $P_f$ is computed as
\begin{equation}
  P_f = \Phi(-\beta),
  \label{eq:pf}
\end{equation}
where $\Phi(\cdot)$ is the cumulative distribution function (CDF) of the standard normal distribution. The above formulation is then inverted to obtain the allowable design variable in the RBD framework. For a prescribed target reliability index $\bbeta$, the objective is to determine the allowable value of the DV, $\lambda$, such that the computed $\beta$ matches the target value. This inverse reliability-based design problem is expressed as
\begin{equation}
  \text{Determine } \lambda \text{ such that } \beta = \bbeta.
  \label{eq:inverse}
\end{equation}

This inverse problem involves an additional optimization process, separate from the forward reliability analysis used to compute $\beta$. In the present study, RBD was carried out for a spherical shell cap made of Al-6061-T6, considering different shell geometries and fabrication quality classes. In practical engineering applications, $\bbeta$ typically varies between 3 and 5; however, values from 1 to 5 are considered here to demonstrate the variation of the allowable DVs and safety-consistent KDFs. The GPR--XGB metamodel is used to represent the actual KDF response, while the design KDF is obtained from code-specified provisions.

\begin{table}[!htbp]
  \centering
  \caption{Safety-consistent KDFs (mean $\mu$ and $\pm1\sigma$ interval bounds) for varying target reliability ($\beta = \bbeta$) for the Al-6061-T6 spherical shell cap ($R/t = 400$, $H/r_0 = 0.58$, $Q = 25$, Class~B), obtained with the hybrid GPR--XGB and ANN metamodels.}
  \label{tab:rbd}
  \begin{tabular}{ccccc}
    \toprule
    $\bbeta$ & $Q$ (Class B) & $\KDF(\mu)$ & $\KDF(\mu+\sigma)$ & $\KDF(\mu-\sigma)$ \\
    \midrule
    \multicolumn{5}{l}{\textit{GPR--XGB}} \\
    1 & \multirow{5}{*}{25} & 0.60 & 0.61 & 0.59 \\
    2 &                     & 0.63 & 0.65 & 0.61 \\
    3 &                     & 0.68 & 0.70 & 0.66 \\
    4 &                     & 0.71 & 0.73 & 0.69 \\
    5 &                     & 0.75 & 0.77 & 0.69 \\
    \midrule
    \multicolumn{5}{l}{\textit{ANN}} \\
    1 & \multirow{5}{*}{25} & 0.55 & 0.58 & 0.52 \\
    2 &                     & 0.58 & 0.61 & 0.55 \\
    3 &                     & 0.62 & 0.65 & 0.59 \\
    4 &                     & 0.65 & 0.69 & 0.63 \\
    5 &                     & 0.68 & 0.73 & 0.65 \\
    \bottomrule
  \end{tabular}
\end{table}

\begin{figure}[!htbp]
  \centering
  \includegraphics[width=0.8\textwidth]{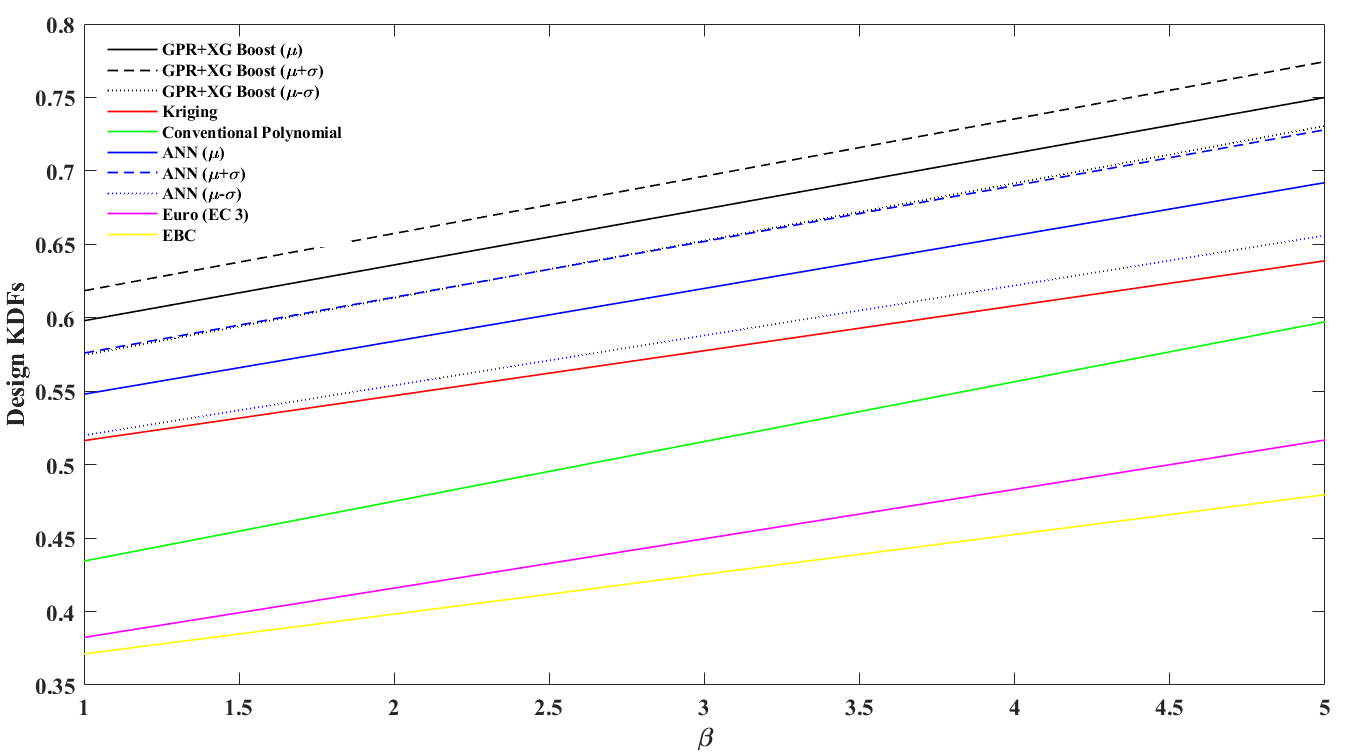}
  \caption{Comparison of GPR--XGB-based safety-consistent design KDFs with other models for varying $\bbeta$ for the Al-6061-T6 spherical shell cap ($R/t = 400$, $H/r_0 = 0.58$, $Q = 25$).}
  \label{fig:rbd}
\end{figure}

\Cref{tab:rbd} indicates that, through the RBD approach, the KDFs can be selected to achieve a desired $\bbeta$, unlike the code-specified KDFs, which depend solely on the geometric parameters of the shell. \Cref{fig:rbd} compares the KDFs derived from the RBD with those prescribed by the design code and other conventional response surfaces (RS) for different $\bbeta$. The comparison clearly reveals that the ML-based KDFs predicted by GPR--XGB are significantly less conservative and less uncertain than the others while ensuring the same level of reliability.

\Cref{fig:rbd} compares the safety-consistent KDFs predicted by the proposed GPR--XGB model, ANN, Kriging~\citep{haldar2000probability}, the conventional polynomial model, Eurocode~3 (EC3)~\citep{rotter2013buckling}, and the Energy Barrier Criterion (EBC)~\citep{wagner2018robust,majumder2024reliability} over the range of the target reliability index $\bbeta$ from 1 to 5. The proposed GPR--XGB model consistently predicts higher design KDFs than the ANN and Kriging models, while maintaining a relatively narrow uncertainty band ($\mu \pm \sigma$), indicating greater prediction confidence and robustness. The ANN exhibits lower mean KDFs and a wider uncertainty interval, suggesting comparatively larger prediction variability. The prediction interval of the GPR--XGB model remains approximately 0.04 over the majority of the investigated reliability indices, whereas the ANN prediction interval is approximately 0.06 over the corresponding range. The narrower prediction band indicates a more consistent surrogate response and reduced variability in the estimated safety-consistent KDFs, thereby increasing confidence in the resulting reliability assessment.

\subsection{Advantages and Novel Contributions of the Proposed GPR--XGB-Assisted RBD Framework}
\label{sec:advantages}

It is observed that the conventional RS underestimates the reliability, whereas the GPR--XGB-based RS reveals a greater safety margin, as shown in \cref{fig:rbd}. The computational efficiency of the proposed hybrid GPR--XGB metamodel stems from replacing computationally expensive nonlinear FE buckling analyses with an accurate data-driven surrogate during reliability analysis and optimization. Once trained, the metamodel predicts the KDF almost instantaneously, allowing repeated evaluations of the LSF without additional nonlinear Riks analyses. Compared with conventional polynomial and ordinary Kriging RS, the proposed framework more accurately captures the highly nonlinear and imperfection-sensitive buckling behaviour through a combination of probabilistic global trend modelling and localized residual learning. The resulting reduction in surrogate error decreases the number of reliability optimization iterations, minimizes the need for additional design of experiments (DOE) refinement, and enables efficient large-scale Monte Carlo or reliability simulations.

Unlike conventional surrogate-based metamodels that rely on predefined basis functions or smooth interpolation assumptions, the proposed GPR--XGB metamodel learns the nonlinear relationship between shell geometry, imperfections, and KDFs directly from experimental data. The hybrid architecture combines the probabilistic global trend captured by GPR with the localized nonlinear residual correction provided by XGBoost. This results in a more faithful approximation of the LSF, reduced surrogate error, improved robustness to noisy and sparse datasets, and enhanced reliability estimation. Consequently, the proposed ML-assisted RBD framework offers greater accuracy, computational efficiency, and adaptability than conventional surrogate-based RBD approaches.

The proposed framework represents a significant advancement over previous studies on KDF prediction and RBD of shell structures. In comparison with the deterministic design methodology of \citet{wagner2018robust}, which developed mechanics-based KDF equations using the EBC, the present study adopts a fully data-driven predictive framework while preserving the RBD philosophy. Unlike the authors' earlier RBD study of spherical shell caps, which relied on FE-based metamodels for repeated evaluations of the LSF considering uncertain-but-bounded KDFs~\citep{majumder2024reliability}, and in comparison with the works of \citet{bazzucchi2017instability} and \citet{cardoso2026basis}, the proposed approach employs ML-based surrogate models, thereby substantially reducing computational effort while providing faster, more flexible, and scalable reliability assessments. Furthermore, compared with the previous ANN-based study of truncated conical shells, which utilized a single ANN model for KDF prediction~\citep{majumder2024predicting}, the physics-informed neural network surrogate developed for reliability-based design of truncated conical shells~\citep{dutta2026physics}, and the hybrid GPR--XGB framework previously used to quantify buckling strength reduction in truncated conical shells~\citep{majumder2026data}, and the ML-based robust prediction and sensitivity assessment of KDFs in truncated conical shells~\citep{majumder2026robust}, the present work extends the methodology to spherical shell caps through a comprehensive comparative evaluation of ML regressors, including ANN, SVR, RFR, HGBoost, and the novel hybrid GPR--XGB framework. In addition, the study incorporates a feature sensitivity analysis to identify the governing parameters influencing the KDFs and integrates the best-performing predictive model within an RBD framework. Consequently, the proposed methodology provides a unified data-driven framework that combines advanced ML, hybrid probabilistic ensemble modelling, sensitivity analysis, and RBD for spherical shell buckling. By moving beyond deterministic formulations, conventional FE-based metamodels, and single-model ML approaches, the proposed framework offers a more accurate, computationally efficient, and practically applicable strategy for the prediction of KDFs and the reliability-consistent design of spherical shell caps.

\section{Conclusion}
\label{sec:conclusion}

The present study developed a comprehensive data-driven framework for predicting the KDFs of imperfection-sensitive spherical shell caps and integrating these predictions within an RBD methodology. Five supervised ML regression models, namely ANN, SVR, RFR, HGBoost, and a novel hybrid GPR--XGB framework, were trained and evaluated using an experimental database comprising 582 spherical shell buckling observations. Among the investigated models, the proposed hybrid GPR--XGB framework demonstrated the best overall predictive performance, achieving an $\Rsq$ of 0.93, an RMSE of 0.056, and an empirical 95\% CI coverage of 96.7\% on the independent test dataset. By combining the probabilistic global trend modelling capability of GPR with the nonlinear residual learning capability of XGBoost, the proposed framework effectively captured the complex imperfection-sensitive buckling behaviour while simultaneously providing calibrated uncertainty estimates and improved robustness against sparse and noisy experimental data.

The study further demonstrated that the proposed hybrid model consistently outperformed the ANN, SVR, RFR, HGBoost, polynomial RSM, and Kriging-based surrogate models in terms of prediction accuracy, uncertainty calibration, and robustness. Calibration analyses, uncertainty-band assessment, and sparsity-based extrapolation studies confirmed that the hybrid model provides reliable confidence estimates while maintaining stable predictive performance throughout the experimentally represented feature space. SHAP-based interpretation additionally identified the fabrication tolerance quality parameter $Q$ and $\lambda$ as dominant variables governing KDF prediction, thereby providing physically meaningful insight into the mechanics of imperfection-sensitive shell buckling.

The developed hybrid predictor was subsequently incorporated into an ML-assisted RBD framework for repeated evaluation of the LSF. Compared with conventional nonlinear FE or polynomial/Kriging surrogate-based RBD procedures, the proposed methodology substantially reduces computational effort while maintaining high predictive fidelity, thereby enabling efficient reliability assessment and design optimization. The framework directly provides safety-consistent KDFs corresponding to specified target reliability indices, replacing the overly conservative deterministic design philosophy adopted in current standards such as NASA SP-8032 and Eurocode~3. The ML-predicted KDFs remained within approximately $\pm5\%$ of the experimentally observed values, whereas the conventional design provisions underestimated the KDFs by approximately 20--40\%, demonstrating the potential for safer and more economical shell designs without compromising structural reliability.

Overall, the proposed hybrid GPR--XGB-assisted reliability framework represents a significant advancement over conventional analytical formulations, classical surrogate models, and single-model ML approaches for spherical shell buckling. The methodology combines predictive accuracy, calibrated uncertainty quantification, computational efficiency, and physical interpretability within a unified design framework, making it a practical decision-support tool for reliability-consistent shell design. Owing to its general data-driven formulation, the framework can be readily extended to other imperfection-sensitive shell geometries, composite shell structures, and reliability-based structural optimization problems where rapid and accurate surrogate modelling is essential.

\FloatBarrier
\begin{appendices}
\crefalias{section}{appendix}\crefalias{subsection}{subappendix}
\appendixpage
\noindent The following appendices provide supporting details: \cref{app:fe} summarizes the linear and nonlinear FE buckling analyses of the shell cap, and \cref{app:meta} outlines the Kriging (GPR) and polynomial response surface metamodels used as baselines in \cref{sec:comparison}.
\numberwithin{equation}{section}
\renewcommand{\theequation}{\thesection.\arabic{equation}}
\renewcommand{\thefigure}{\thesection.\arabic{figure}}
\renewcommand{\thetable}{\thesection.\arabic{table}}
\setcounter{figure}{0}
\setcounter{table}{0}

\section{Linear and Nonlinear Buckling Analysis of the Shell Cap}
\label{app:fe}

\subsection{Linear Buckling Analysis}
\label{app:linear}

\begin{figure}[!htbp]
  \centering
  \includegraphics[width=0.8\textwidth]{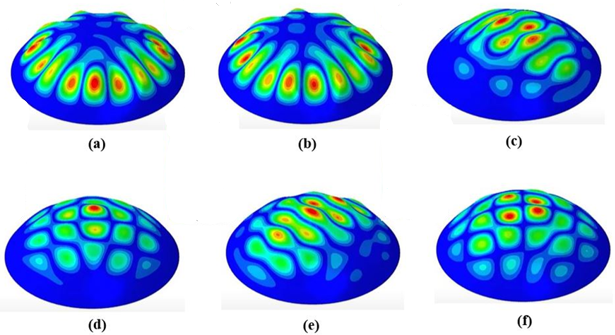}
  \caption{(a)--(f) First six linearized buckling modes from eigen-analysis \citep{majumder2024reliability}.}
  \label{fig:eigenmodes}
\end{figure}

In linear buckling analysis, the buckling problem is simplified into an eigenvalue formulation, where the eigenvalues define the critical pressures (loads) and the eigenvectors represent the associated buckling modes. For spherical shell caps, linearized eigenvalue analysis provides the initial estimates. For the shell geometry adopted in \cref{sec:fe}, the first six eigenmodes are shown in \cref{fig:eigenmodes}(a)--(f). A fundamental critical pressure of 0.527~MPa was obtained, which closely matches the value of 0.53~MPa predicted by \cref{eq:zoelly}. These values correspond to a perfect shell cap.

\subsection{Nonlinear Stability (Buckling) Analysis}
\label{app:nonlinear}

In practice, the critical loads of spherical shells are far lower than those predicted by linear analysis, which necessitates nonlinear FE methods. Here, the incremental--iterative Riks method is applied, enabling combined displacement and load control while capturing snap-through and snap-back responses along the nonlinear load--displacement path. The analysis begins with an initial compressive load, typically the first critical load from the eigen-analysis, and iteratively yields a load proportionality factor (LPF). Multiplying the LPF by the initial load gives the actual shell load at any deformation stage, while the peak LPF identifies the KDF.

\begin{figure}[!htbp]
  \centering
  \includegraphics[width=\textwidth]{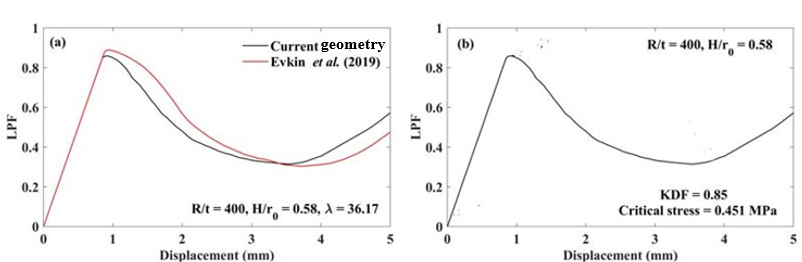}
  \caption{(a) Equilibrium load--displacement path of the perfect shell cap under uniform external pressure and (b) the pertinent LPFs at various instants of deformation \citep{majumder2024reliability}.}
  \label{fig:riks}
\end{figure}

The normalized stress--displacement curves reveal behaviour consistent with prior studies, such as \citet{evkin2019design}. The path first rises to a pressure peak and then descends with further deflection, highlighting the instability typical of shells of revolution. Snap-through responses occur if load control is enforced, while displacement control captures the stable continuation of the equilibrium path. The LPFs at various configurations demonstrate significant post-critical reduction, with the fundamental stress dropping by about 15\%, as shown in \cref{fig:riks}(a)--(b) for a perfect shell cap devoid of imperfections.

The deformed shell configuration is shown in the authors' previous study~\citep{majumder2024reliability}. The post-critical load reduction is expected to become larger with increasing imperfections.

\section{Kriging and Polynomial Response Surface Metamodels}
\label{app:meta}

\subsection{Gaussian Process Regression (Kriging)}
\label{app:gpr}

GPR, also referred to as Kriging, is implemented in this study to model the relationship between the standardized input feature vector $\x$ and the KDF response. A Gaussian process defines a prior over possible response functions, where the function values are assumed to follow a joint Gaussian distribution specified by a mean function and a covariance kernel:
\begin{equation}
  f(\x) \sim \GP\bigl(m(\x),\, k(\x,\x')\bigr),
  \label{eq:gp_prior}
\end{equation}
where $m(\x)$ is the mean function and $k(\x,\x')$ is the covariance function. After standardizing the KDF observations, the mean function is taken as zero. Conditioning on the training inputs $\X$ and target KDFs $\y$, the posterior predictive distribution of the latent function value $f_{*}$ at a test input $\x_{*}$ is Gaussian:
\begin{equation}
  f_{*} \mid \X, \y, \x_{*} \sim \Normal\bigl(\mu_{*},\, \sigma_{*}^{2}\bigr).
  \label{eq:gp_post}
\end{equation}
The corresponding predictive mean is given by
\begin{equation}
  \mu_{*} = \kv_{*}^{\T}\bigl(\K + \sigma_n^{2}\I\bigr)^{-1}\y,
  \label{eq:gp_mean}
\end{equation}
and the predictive variance is
\begin{equation}
  \sigma_{*}^{2} = k(\x_{*},\x_{*}) - \kv_{*}^{\T}\bigl(\K + \sigma_n^{2}\I\bigr)^{-1}\kv_{*}.
  \label{eq:gp_var}
\end{equation}
Here, $\K$ is the $n \times n$ training covariance (Gram) matrix with entries
\begin{equation}
  K_{ij} = k(\x_i,\x_j).
  \label{eq:gram}
\end{equation}
The vector $\kv_{*}$ contains the covariances between the test input $\x_{*}$ and all training inputs, $\I$ is the identity matrix, and $\sigma_n^{2}$ is the variance of the additive observation noise. The predicted KDF is represented by $\mu_{*}$, while $\sigma_{*}$ provides the corresponding predictive uncertainty.

The covariance function used in this study is a composite additive kernel consisting of a Mat\'ern kernel, a radial basis function (RBF) kernel, and a white-noise kernel:
\begin{equation}
  k(\x,\x') = k_{\mathrm{Mat\acute{e}rn}}(\x,\x') + k_{\mathrm{RBF}}(\x,\x') + k_{\mathrm{White}}(\x,\x').
  \label{eq:kernel_sum}
\end{equation}
The kernel is initialized as
\begin{equation*}
  k(\x,\x') = C(1.0)\bigl[k_{\mathrm{Mat\acute{e}rn}}(l = 1.0,\ \nu = 2.5) + k_{\mathrm{RBF}}(l = 0.5)\bigr] + k_{\mathrm{White}}.
\end{equation*}
The general Mat\'ern covariance function is given by
\begin{equation}
  k_{\mathrm{Mat\acute{e}rn}}(d) = \sigma_f^{2}\,\frac{2^{1-\nu}}{\Gamma(\nu)}\left(\frac{\sqrt{2\nu}\,d}{l}\right)^{\nu} K_{\nu}\!\left(\frac{\sqrt{2\nu}\,d}{l}\right),
  \label{eq:matern}
\end{equation}
where the Euclidean distance between two input points is
\begin{equation}
  d = \lVert \x - \x' \rVert.
  \label{eq:distance}
\end{equation}
In this expression, $l$ is the characteristic length scale, $\sigma_f^{2}$ is the signal variance, $\Gamma(\cdot)$ is the gamma function, and $K_{\nu}(\cdot)$ is the modified Bessel function of the second kind. This kernel produces once-differentiable sample paths and is therefore suitable for capturing localized and moderately non-smooth variations associated with imperfection sensitivity in the KDF response.
The RBF component captures the smoother, globally varying trend and is expressed as
\begin{equation}
  k_{\mathrm{RBF}}(d) = \sigma_f^{2}\exp\left(-\frac{d^{2}}{2l^{2}}\right).
  \label{eq:rbf}
\end{equation}
The white-noise kernel accounts for irreducible measurement noise and unresolved experimental variability:
\begin{equation}
  k_{\mathrm{White}}(\x,\x') = \sigma_n^{2}\,\delta(\x,\x'),
  \label{eq:white}
\end{equation}
where $\delta(\x,\x')$ is the Kronecker delta. The hyperparameters of each kernel component, $\bm{\theta} = \{\sigma_f^{2},\, l,\, \sigma_n^{2}\}$, are re-optimized within every training fold by maximizing the log marginal likelihood:
\begin{equation}
  \log p(\y \mid \X, \bm{\theta}) = -\frac{1}{2}\y^{\T}\bigl(\K_{\theta} + \sigma_n^{2}\I\bigr)^{-1}\y - \frac{1}{2}\log\bigl\lvert \K_{\theta} + \sigma_n^{2}\I \bigr\rvert - \frac{n}{2}\log(2\pi).
  \label{eq:lml}
\end{equation}
The optimization is performed with multiple random restarts to reduce the likelihood of convergence to a poor local optimum.

\subsection{Polynomial Response Surface Metamodel (RSM)}
\label{app:rsm}

The RSM is used as a deterministic baseline regression model to approximate the relationship between the input features and the target KDF response. Unlike Kriging (GPR), which treats the unknown response as a stochastic function, the RS approach assumes that the response can be represented explicitly by a low-order polynomial function of the input variables. In this study, a second-order polynomial RS is adopted, together with input standardization and ridge regularization. The original input feature vector $\x$ is converted to a standardized vector $\z$. The quadratic response surface is constructed by augmenting the standardized input vector with all linear terms, squared terms, and pairwise interaction terms. The general second-order response surface model is expressed as
\begin{equation}
  \hat{y} = \beta_0 + \sum_{j=1}^{d}\beta_j z_j + \sum_{j=1}^{d}\beta_{jj} z_j^{2} + \sum_{j=1}^{d-1}\sum_{k=j+1}^{d}\beta_{jk} z_j z_k,
  \label{eq:rsm}
\end{equation}
where $\hat{y}$ is the predicted KDF, $\beta_0$ is the intercept, $\beta_j$ are the coefficients of the linear terms, $\beta_{jj}$ are the coefficients of the quadratic terms, and $\beta_{jk}$ are the coefficients of the pairwise interaction terms. The linear terms represent the first-order contribution of each input feature, the squared terms capture curvature in the response, and the interaction terms account for coupled effects between pairs of variables. For compact notation, the polynomial feature vector may be written as
\begin{equation}
  \bphi(\z) = \bigl[z_1, \ldots, z_d,\ z_1^{2}, \ldots, z_d^{2},\ z_1 z_2,\ z_1 z_3, \ldots, z_{d-1} z_d\bigr]^{\T}.
  \label{eq:phi}
\end{equation}
Using this polynomial basis, the RS prediction can be written in vector form as
\begin{equation}
  \hat{y} = \beta_0 + \bphi(\z)^{\T}\bbet,
  \label{eq:rsm_vec}
\end{equation}
where $\bbet$ is the vector of regression coefficients associated with the polynomial basis terms. For a training dataset containing $n$ samples, the polynomial feature matrix is defined as
\begin{equation}
  \bPhi = \begin{bmatrix}
    \bphi(\z_1)^{\T} \\
    \bphi(\z_2)^{\T} \\
    \vdots \\
    \bphi(\z_n)^{\T}
  \end{bmatrix}.
  \label{eq:Phi}
\end{equation}
In the present implementation, the polynomial coefficients are estimated using ridge regression rather than ordinary least squares. Ridge regression introduces an $L_2$-regularization penalty to control the magnitude of the regression coefficients and improve numerical stability, particularly when the polynomial basis contains correlated terms. The coefficient estimation problem is written as
\begin{equation*}
  \min_{\beta_0,\,\bbet}\left[\sum_{i=1}^{n}\bigl(y_i - \beta_0 - \bphi(\z_i)^{\T}\bbet\bigr)^{2} + \alpha\,\bbet^{\T}\bbet\right],
\end{equation*}
where $\alpha$ is the ridge regularization parameter. In this study, $\alpha = 1.0$ was used. The intercept $\beta_0$ was not penalized, while the remaining polynomial coefficients were regularized. In matrix form, the ridge regression solution can be expressed as
\begin{equation}
  \bbet = \bigl(\bPhi^{\T}\bPhi + \alpha\I\bigr)^{-1}\bPhi^{\T}\y,
  \label{eq:ridge_sol}
\end{equation}
where $\I$ is the identity matrix. When $\alpha = 0$, the formulation reduces to the ordinary least-squares response surface model. A positive regularization parameter helps reduce overfitting by discouraging excessively large polynomial coefficients. For a new test input $\x_{*}$, the same standardization transformation is first applied to obtain the standardized test vector $\z_{*}$, which is then expanded into the same quadratic polynomial basis $\bphi(\z_{*})$; the predicted KDF is obtained as
\begin{equation}
  \hat{y}_{*} = \beta_0 + \bphi(\z_{*})^{\T}\bbet.
  \label{eq:rsm_pred}
\end{equation}
Thus, the RSM provides a deterministic polynomial approximation of the KDF response. The quadratic terms allow the model to represent smooth nonlinear variations, while the interaction terms allow the influence of one variable to depend on the value of another. However, because the response is restricted to a second-order polynomial form, the model may not fully capture highly nonlinear, localized, or discontinuous variations in the KDF response. In addition, unlike Kriging, the response surface model does not directly provide predictive uncertainty unless an additional residual-error model is introduced.
\end{appendices}

\FloatBarrier
\section*{Acknowledgements}
The authors acknowledge the SEED Grant (Ref.\ No.\ AU/SRG/2025-01) received from Adani University, Ahmedabad, Gujarat, India, in support of this research.

\bibliography{references}

\end{document}